%% file: main.tex
\documentclass[10pt,twocolumn,letterpaper]{article}

\usepackage[pagenumbers]{cvpr} 
\usepackage[accsupp]{axessibility}

\input{preamble}

\definecolor{cvprblue}{rgb}{0.21,0.49,0.74}
\usepackage[pagebackref,breaklinks,colorlinks,allcolors=cvprblue]{hyperref}

\def\paperID{***} 
\def\confName{CVPR}
\def\confYear{2026}

\title{
\raisebox{-0.35\height}{\includegraphics[height=1.8em]{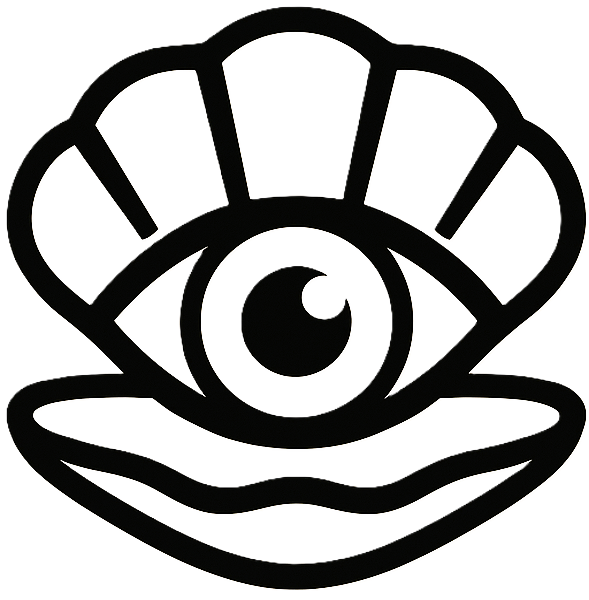}}\hspace{0.3em}%
PEARL: Geometry Aligns Semantics for \\ Training-Free Open-Vocabulary Semantic Segmentation
}

\author{Gensheng Pei$^{1}$, Xiruo Jiang$^{2}$, Xinhao Cai$^{3}$, Tao Chen$^{3}$, Yazhou Yao$^{3}$, Byeungwoo Jeon$^{1}$\thanks{Corresponding author.} \\
\small{$^{1}$Department of Electrical and Computer Engineering, Sungkyunkwan University} \\
\small{$^{2}$School of Computing and Artificial Intelligence, Southwest Jiaotong University} \\
\small{$^{3}$School of Computer Science and Engineering, Nanjing University of Science and Technology} \\
\small{\url{https://github.com/PGSmall/PEARL}} \\
}

\begin{document}
\maketitle
\input{sec/0_abstract}
\input{sec/1_introduction}
\input{sec/2_related_work}
\input{sec/3_method}

\input{sec/4_experiments}
\input{sec/5_conclusion}

{
    \small
    \bibliographystyle{ieeenat_fullname}
    \bibliography{main}
}

\input{sec/6_appendix}

\end{document}

%% file: preamble.tex
\usepackage{booktabs, makecell, multirow, xcolor, colortbl}
\usepackage{pifont}
\usepackage{bm}
\usepackage{graphicx}
\usepackage{tikz}
\usetikzlibrary{positioning,fit,calc,backgrounds}

\definecolor{Best}{RGB}{242,249,238}
\definecolor{Second}{RGB}{239,246,255}
\definecolor{highlight}{RGB}{30, 144, 255}

\newcommand{\cmark}{\textcolor{green!50!black}{\ding{51}}}
\newcommand{\xmark}{\textcolor{red!70!black}{\ding{55}}}

\newcommand{\best}[1]{\cellcolor{Best}\textbf{#1}}
\newcommand{\second}[1]{\cellcolor{Second}\underline{#1}}
\newcommand{\up}[1]{\textcolor{green!50!black}{\textbf{#1}}}



%% file: sec/0_abstract.tex
\begin{abstract}
Training-free open-vocabulary semantic segmentation (OVSS) promises rapid adaptation to new label sets without retraining. Yet, many methods rely on heavy post-processing or handle text and vision in isolation, leaving cross-modal geometry underutilized. Others introduce auxiliary vision backbones or multi-model pipelines, which increase complexity and latency while compromising design simplicity.
We present PEARL, \textbf{\underline{P}}rocrust\textbf{\underline{e}}s \textbf{\underline{a}}lignment with text-awa\textbf{\underline{r}}e \textbf{\underline{L}}aplacian propagation, a compact two-step inference that follows an align-then-propagate principle. The Procrustes alignment step performs an orthogonal projection inside the last self-attention block, rotating keys toward the query subspace via a stable polar iteration. The text-aware Laplacian propagation then refines per-pixel logits on a small grid through a confidence-weighted, text-guided graph solve: text provides both a data-trust signal and neighbor gating, while image gradients preserve boundaries. In this work, our method is fully training-free, plug-and-play, and uses only fixed constants, adding minimal latency with a small per-head projection and a few conjugate-gradient steps. Our approach, PEARL, sets a new state-of-the-art in training-free OVSS without extra data or auxiliary backbones across standard benchmarks, achieving superior performance under both with-background and without-background protocols.
\end{abstract}

%% file: sec/1_introduction.tex
\section{Introduction}
\label{sec:intro}

\begin{figure}
    \centering
    \includegraphics[width=1\linewidth]{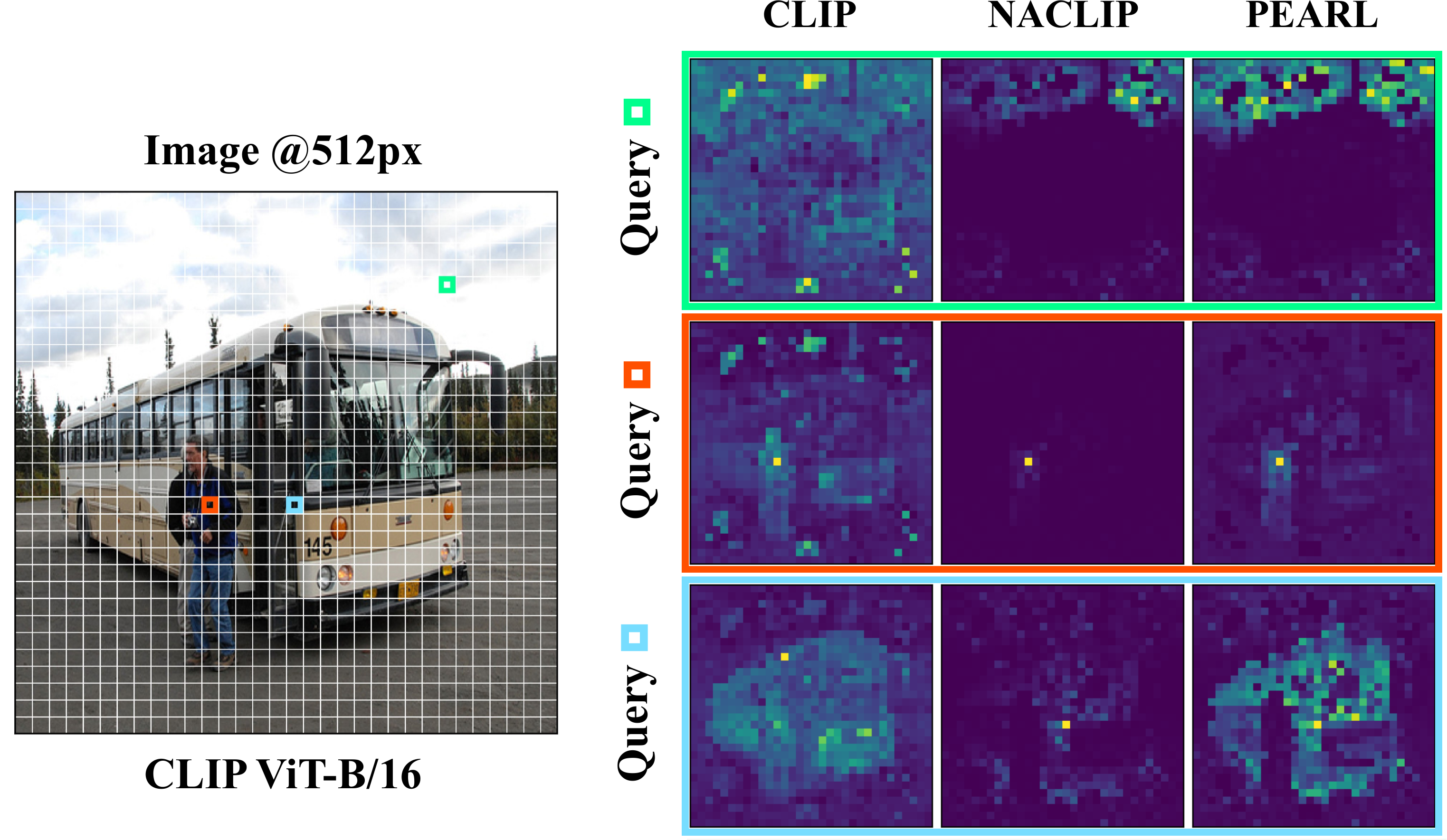}
    \vspace{-0.65cm}
    \caption{\textbf{Attention visualization at selected query points.}
    We show an input resized to $512\times512$ using the CLIP ViT-B/16 vision encoder.
    For each colored query, we compare the attention maps produced by CLIP~\cite{clip}, NACLIP~\cite{naclip}, and our PEARL.
    CLIP exhibits diffuse and background-biased responses, while NACLIP improves localization but often fragments objects into isolated peaks.
    By performing Procrustes alignment in the last self-attention block, PEARL yields compact and object-consistent attention, suppresses background spill, and better preserves thin parts, resulting in stable focus with reduced grid artifacts.}
    \label{fig:attn}
    \vspace{-0.5cm}
\end{figure}

Open-vocabulary semantic segmentation~\cite{ovsp,maskclip,ovs_maclip,zsseg,sed,ovs_pacl,sanet_ovss} (OVSS) assigns a category to every pixel when the label set is specified at inference time through natural language. In the training-free setting, a frozen vision-language backbone provides dense visual features and text embeddings, and masks are obtained by matching patches to text prototypes without any task-specific optimization~\cite{talk2dino,cass}. Early progress \cite{ovsp,zero-shot-semseg} built on zero-shot scene parsing \cite{hcpn,hgpu} with word embeddings. The field accelerated with contrastive vision-language models (VLMs) such as CLIP~\cite{clip} and ALIGN~\cite{align}, paired with strong Transformer backbones like ViT~\cite{ViT,deit,swin} and the DINO family~\cite{dinov1,dinov2,dinov3}. Supervised segmentation remains a touchstone for dense prediction quality, but OVSS aims to retain flexibility while avoiding retraining on every new label set.

Training-based approaches~\cite{zsseg,sed,ovs_maclip,cat-seg} convert these ingredients into dense predictors by learning decoders, adding lightweight adapters, or leveraging weak supervision. Representative work attaches decoders to frozen or partially tuned backbones~\cite{zsseg,lseg,sed}, adapts CLIP with mask-aware objectives or side modules~\cite{ovs_maclip,sanet_ovss,cat-seg}, and exploits image labels, boxes, or captions to reduce mask annotation~\cite{openseg,detic,regionclip,glip,groupvit,tcl,ovs_nls,ovps_diff}. A complementary line distills object-centric structure from self-supervised ViTs into CLIP-like architectures to inject stronger grouping cues~\cite{dinoisers,talk2dino,dino.txt}. These routes can reach high accuracy, but require extra data and task-specific training.

In parallel, the training-free paradigm keeps backbones frozen, improving inference only. A pioneering baseline, MaskCLIP~\cite{maskclip}, computes cosine similarity between dense CLIP features and text prompts. Subsequent work strengthens this recipe along three directions. First, feature purification and re-alignment suppress outliers, rebuild patch correlations, or select informative attention heads~\cite{sfp,clearclip,cliptrase,corrclip,reme,gem}. Second, spatial refinement encourages coherence using classical mask refinement and neighbor-aware grouping~\cite{densecrf,pamr,clip-diy,CaR,naclip}. Third, object and context priors are imported through spectral cues or multi-model assemblies, \eg, distilling DINO-style structure~\cite{lposs,ProxyCLIP,lavg} or leveraging visual context graphs~\cite{cass}. These approaches can be practical, though heavy post-processing or auxiliary components may increase complexity and latency.

Two observations motivate our work. (\textit{i}) Contrastive pretraining emphasizes global image-text agreement rather than dense prediction. Near the top of the vision encoder, a few background-dominated directions can steer token interactions, yielding patch geometry that is misaligned and spatially inconsistent for pixel-level decisions~\cite{maskclip,clearclip}. When the geometry at the source is off, downstream smoothing treats symptoms rather than causes. (\textit{ii}) Text is commonly used only as a classifier. It rarely governs how pixels exchange information, even though relations in the text space suggest which categories should reinforce one another and which should remain separate~\cite{naclip,cass}. These two points suggest a simple strategy: first correct the geometry where attention scores are formed, then propagate semantics with guidance from both text relations and image boundaries.

Following this strategy, we present PEARL, a \textbf{P}rocrust\textbf{e}s \textbf{a}lignment with text-awa\textbf{r}e \textbf{L}aplacian propagation. The first step inserts an orthogonal Procrustes alignment into the last self-attention block. After weighted centering of queries and keys, a single input-dependent rotation aligns keys to the query subspace. This closed-form correction preserves magnitudes and angles while removing background-biased drift that destabilizes patch-text similarities.
As shown in Fig.~\ref{fig:attn}, attention maps at selected queries (using CLIP ViT-B/16) show that vanilla CLIP is diffuse and background-biased, while NACLIP~\cite{naclip} sharpens focus but fragments objects, our PEARL yields compact, object-consistent responses that preserve thin structures. This suggests that aligning keys to queries before scoring stabilizes token geometry and leads to cleaner masks.
The second step refines the class-logit field on a compact grid using a text-aware Laplacian: a confidence-weighted data term trusts pixels with reliable evidence, while a text agreement prior gates neighbor links according to semantic relatedness. Image gradients protect boundaries, ensuring that refinement respects edges~\cite{densecrf,pamr}. Unlike class-agnostic smoothing, text guides both the data trust and the edges. Unlike heavy spectral or multi-backbone assemblies, the refinement is a single linear solve with modest cost~\cite{cass}.

To summarize, our PEARL addresses the two weaknesses above at their origin with two principled operators. Procrustes alignment repairs token geometry exactly where attention is computed, and text-aware Laplacian propagation transforms language from a simple labeler into a structural prior that guides how evidence spreads across the image. The two parts work together: the first enhances what attention sees, and the second organizes how pixels align. Across standard OVSS benchmarks, the approach achieves coherent masks for small objects, stable coverage for ``\textit{stuff}" regions, and strong accuracy among training-free methods, all without auxiliary backbones or extra supervision.

%% file: sec/2_related_work.tex
\section{Related Work}
\label{sec:related_work}

\noindent\textbf{From Closed-Set to Open-Vocabulary Semantic Segmentation.}
Classical semantic segmentation~\cite{fcn, deeplab, xie21_segformer}, built on pixel-annotated datasets (\eg, PASCAL VOC~\cite{pascal_voc}, MS COCO~\cite{coco}, ADE20K~\cite{ade, ade_ijcv}, Cityscapes~\cite{Cityscapes}), works well but is limited to a \textit{closed set} of categories. Extending them to new concepts requires costly re-annotation and re-training. To overcome this, the field has shifted towards OVSS. Early steps toward open settings used word embeddings for zero-shot transfer~\cite{ovsp, zero-shot-semseg}. The move to open-vocabulary segmentation accelerated with the emergence of VLMs like CLIP~\cite{clip} and ALIGN~\cite{align}, which align images and text in a shared space. Yet CLIP is trained for global alignment, not dense prediction: patch features are often noisy and spatially imprecise. Most recent work, therefore, aims to recover spatially coherent, text-aware masks either by adapting the model with additional data or by reshaping the inference process.
We take the latter route with our method, PEARL: a training-free, plug-and-play \textit{align-then-propagate} scheme that first corrects \textit{feature geometry} and then performs \textit{text-aware propagation} entirely at inference.

\noindent\textbf{Training-Based OVSS.}
This family adapts pre-trained VLMs~\cite{clip, align, eva, clip_pgs} or VMs~\cite{dinov1, dinov2, sam, samv2} to produce dense masks, using additional data to fine-tune a decoder, add small adapters, or train with weak labels. Many methods reduce the gap to dense prediction by learning on task data. Some attach a decoder to frozen or partially tuned VLMs~\cite{zsseg, lseg, sed}, others adapt CLIP via mask-aware training or lightweight adapters~\cite{ovs_maclip, sanet_ovss, cat-seg}, and a broad line leverages weaker supervision from image labels, boxes, or captions~\cite{openseg, detic, regionclip, glip, groupvit, tcl, ovs_nls, ovps_diff}. A complementary trend fuses foundation models: object-centric signals from self-supervised ViTs (\eg, DINO~\cite{dinov1, dinov2, dinov3}) are distilled or aligned into CLIP~\cite{dinoisers, dino.txt, talk2dino, sam-clip}. These approaches deliver strong accuracy but incur curation costs, training bias toward seen concepts, and additional compute requirements.
By contrast, our proposed approach, PEARL, needs no extra training or data and runs parameter-free inference, combining an \textit{orthogonal Procrustes alignment} in the last self-attention with a \textit{text-aware Laplacian propagation} to keep the design simple while preserving generalization.

\noindent\textbf{Training-Free OVSS.}
Training-free paradigm \textit{does not train} anything new. It keeps backbones frozen and modifies only the inference procedure to obtain text-aware masks. A basic baseline computes cosine similarity between dense CLIP features and text prompts~\cite{maskclip}, then improves it along three directions. First, feature purification and re-alignment suppresses outlier/background tokens, rebuilds patch correlations, or selects informative attention heads~\cite{sfp, freecp, reme, cliptrase, corrclip, dih-clip, kang2025your, gem, conceptbank}, with recent work targeting class redundancy and vision-language ambiguity~\cite{freecp}. Second, spatial refinement with locality priors enforces coherence using mask refinement techniques~\cite{densecrf,pamr}, grouping and count-aware propagation~\cite{clip-diy, CaR, pnp-ovss}, or neighbor/feedback-aware attention mechanism~\cite{naclip, fsa, fossil, freeda, resclip, sc_clip}. Third, object-context fusion imports structure from stronger vision backbones or multi-model pipelines (\eg, DINO~\cite{dinov1}, Diffusion~\cite{diffusion}, SAM~\cite{sam}) for correspondences, prototypes, or prompt-based refinement~\cite{clearclip, ProxyCLIP, freeda, fossil, trident}. Previous methods are effective but often rely on multi-stage heuristics, intensive post-processing, or auxiliary backbones.
In this work, we introduce PEARL, Procrustes alignment with text-aware Laplacian propagation, an \textit{align-then-propagate} framework in which an \textit{orthogonal Procrustes} step fixes attention geometry and a confidence-weighted, \textit{text-guided Laplacian} refines logits on a small grid to yield coherent masks with a fully training-free, plug-and-play pipeline.

%% file: sec/3_method.tex
\section{Method}

\subsection{Preliminaries: Training-free OVSS}
\label{sec:pre}

\noindent\textbf{Vision Encoder.} Training-free open-vocabulary semantic segmentation assigns pixels to natural-language concepts while keeping the vision-language backbone frozen. Let the input image be $\bm{I}\in\mathbb{R}^{H\times W\times 3}$. A ViT vision encoder partitions $\bm{I}$ into non-overlapping patches of size $P\times P$, yielding a grid of $H_p\times W_p$ patches with $H_p=H/P$ and $W_p=W/P$. Denote by $\bm{X}\in\mathbb{R}^{N\times D}$ the token matrix at the last Transformer block, with $N=1+H_pW_p$ including the \texttt{CLS} token, and write $D=J\,d$ for $J$ heads of width $d$.

After layer normalization, the final block applies multi-head self-attention. For head $j\in\{1,\ldots,J\}$, linear projections produce $\bm{Q}^{(j)},\bm{K}^{(j)},\bm{V}^{(j)}\in\mathbb{R}^{N\times d}$, and attention is computed as follows:
\begin{equation}
\label{eq:sa}
\begin{split}
    \bm{A}^{(j)}&=\mathtt{softmax}\!\big(d^{-1/2}\,\bm{Q}^{(j)}(\bm{K}^{(j)})^\top\big),\\
    \bm{Y}^{(j)}&=\bm{A}^{(j)}\bm{V}^{(j)} .
\end{split}
\end{equation}
The head outputs are concatenated and projected, $\bm{X}^{\star}=\mathtt{Concat}\big(\bm{Y}^{(1)},\ldots,\bm{Y}^{(J)}\big)\bm{W}^{o}\in\mathbb{R}^{N\times D}$, and the $H_pW_p$ patch rows of $\bm{X}^{\star}$ are taken as per-patch features $\bm{v}[h,w]\in\mathbb{R}^{D}$ on the grid $(h,w)\in\{1,\ldots,H_p\}\times\{1,\ldots,W_p\}$.

\noindent\textbf{Text Encoder.}
A frozen text encoder maps the label set $\mathcal{Y}=\{y_c\}_{c=1}^{C}$ to unit-norm prototypes $\bm{t}_c\in\mathbb{R}^{D}$, stacked as $\bm{T}=[\bm{t}_1^\top;\ldots;\bm{t}_C^\top]\in\mathbb{R}^{C\times D}$. 
Patch-text matching then forms initial per-class logits on the patch grid by cosine-scaled similarity $\bm{Z}\in\mathbb{R}^{H_p\times W_p\times C}$:
\begin{equation}
\label{eq:pre_logits}
\bm{Z}[h,w,c]
=\alpha\,\frac{\bm{v}[h,w]\cdot \bm{t}_c}{\|\bm{v}[h,w]\|_2\,\|\bm{t}_c\|_2},
\qquad
\alpha=D^{-1/2}.
\end{equation}
When computed in a single pass, $\bm{Z}$ is bilinearly upsampled to image resolution, yielding $\hat{\bm{Z}}\in\mathbb{R}^{H\times W\times C}$.

\noindent\textbf{Inference Stage.}
For high-resolution inputs, the image $\bm{I}$ is covered by $M$ overlapping windows $\{\hat{\bm{I}}_{m}\}_{m=1}^{M}$. 
Each window is processed as in Eqs.~\eqref{eq:sa} and \eqref{eq:pre_logits} to obtain its upsampled logits $\widehat{\bm{Z}}^{(m)}\in\mathbb{R}^{H\times W\times C}$ in the image coordinate frame. 
The full-image logits are obtained by weighted fusion on overlaps and are represented as:
\begin{equation}
\widehat{\bm{Z}}[h,w,c]
=\sum_{m=1}^{M}\omega_{m}(h,w)\,\widehat{\bm{Z}}^{(m)}[h,w,c],
\end{equation}
where $\sum_{m=1}^{M}\omega_{m}(h,w)=1$, and $\omega_{m}(h,w)$ are averaging or application-specific weights that vanish outside window $m$. 
The zero-shot segmentation map is then given by:
\begin{equation}
\label{eq:argmax}
\bm{S}(\bm{I},\mathcal{Y})[h,w]
=\arg\max_{c\in\{1,\ldots,C\}}\,\widehat{\bm{Z}}[h,w,c].
\end{equation}

Eqs. \eqref{eq:sa}-\eqref{eq:argmax} describe a single, continuous pipeline: the last self-attention block generates patch features, language prototypes provide the semantic anchor, patch-text similarities define class scores, and pixel labels are followed by selection. This baseline presumes a compatible geometry between vision tokens and text prototypes at the patch level. In practice, the contrastive pretraining that prioritizes global alignment of the \texttt{CLS} token often leaves patch interactions noisy and spatially inconsistent. The following sections operate exactly at these two points of fragility: the attention computation in Eq. \eqref{eq:sa} and the score field in Eq. \eqref{eq:pre_logits}.

\begin{figure*}[t]
    \centering
    \includegraphics[width=0.99\linewidth]{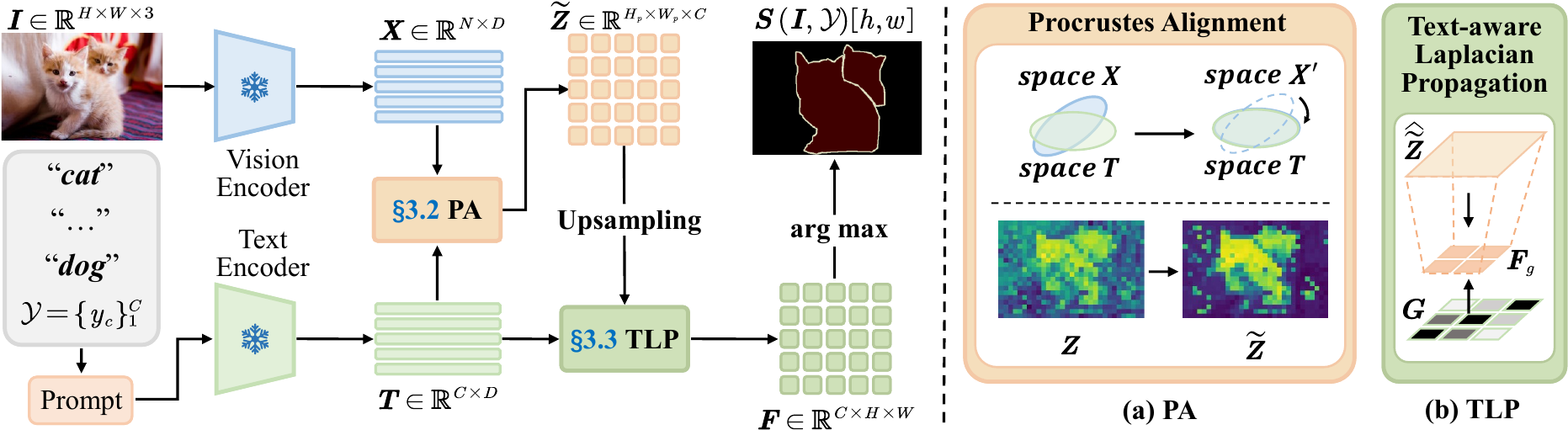}
    \vspace{-0.2cm}
    \caption{
    \textbf{Our PEARL framework with \textit{align-then-propagate}.}
    Given an image $\bm{I}$ and a label set $\mathcal{Y}$, a frozen ViT vision encoder yields patch tokens.
    \emph{Procrustes alignment} is inserted in the last self-attention block: for each head ($J$ heads), queries and keys are weighted-centered and the keys are orthogonally rotated toward the query subspace, producing geometry-corrected patch features.
    A frozen text encoder maps $\mathcal{Y}$ to unit-norm prototypes $\bm{T}$. Patch-text cosine similarity gives initial logits $\widetilde{\bm{Z}}$.
    \emph{Text-aware Laplacian propagation} then refines $\widetilde{\bm{Z}}$ on a compact grid using a class graph $\bm{G}$ induced by $\bm{T}$ and confidence weights, and the refined scores are upsampled to full resolution $\bm{F}$.
    The final segmentation $\bm{S}$ is obtained by selecting, at each pixel, the class with the highest score.}
    \label{fig:framework}
    \vspace{-0.3cm}
\end{figure*}

\subsection{Procrustes Alignment in Self-Attention}
\label{sec:pa}

Patch-text scores in Eq.~\eqref{eq:pre_logits} assume that the patch features $\bm{v}[h,w]$ live in a geometry compatible with the text prototypes. Contrastive pretraining emphasizes global image-text alignment and the last self-attention in Eq.~\eqref{eq:sa} can be dominated by a few background directions, which weakens local correspondence. Before forming scores inside the last block, align the key space to the query space by a single orthogonal map computed per head and per input, as shown in Fig.~\ref{fig:framework}. This preserves norms and angles while correcting the basis mismatch that creates unstable similarities.

Consider one head $j$ (index omitted in matrices for clarity) with queries, keys, and values $\bm{Q},\bm{K},\bm{V}\in\mathbb{R}^{N\times d}$ taken from the last block (please refer to \S\ref{sec:pre}). Tokens are first re-centered to remove global bias. Define nonnegative token weights $\pi_n\propto\|\bm{q}_n\|_2$ with $\sum_{n=1}^{N}\pi_n=1$ and optionally set the \texttt{CLS} weight to zero. With $\bm{1}\in\mathbb{R}^{N\times 1}$, the weighted centroids and centered clouds are computed as follows:
\begin{equation}
\label{eq:center}
\begin{split}
    \bm{\mu}_Q=\sum_{n=1}^{N}\pi_n\bm{q}_n,\qquad
    \bm{\mu}_K=\sum_{n=1}^{N}\pi_n\bm{k}_n,\\
    \bm{Q}_c=\bm{Q}-\bm{1}\bm{\mu}_Q^\top,\qquad
    \bm{K}_c=\bm{K}-\bm{1}\bm{\mu}_K^\top.
\end{split}
\end{equation}
An orthogonal Procrustes problem aligns keys to queries:
\begin{equation}
\label{eq:proc}
\bm{R}^\star=\arg\min_{\bm{R}\in O(d)}\ \|\bm{K}_c\bm{R}-\bm{Q}_c\|_F^2
\Longleftrightarrow
\bm{R}^\star=\bm{U}\bm{V}^\top,
\end{equation}
where $\bm{K}_c^\top\bm{Q}_c=\bm{U}\bm{\Sigma}\bm{V}^\top$. The optimizer is the orthogonal factor of the cross-covariance. It can also be obtained by a few Newton-Schulz iterations on the polar factor for an SVD-free implementation. Only keys are rotated and attention is recomputed within the same block:
\begin{equation}
\label{eq:aligned_attn}
\begin{split}
    \widetilde{\bm{K}}&=\bm{K}\bm{R}^\star, \\
    \widetilde{\bm{A}}&=\mathtt{softmax}\!\big(d^{-1/2}\,\bm{Q}\widetilde{\bm{K}}^\top\big),\\
    \widetilde{\bm{Y}}&=\widetilde{\bm{A}}\bm{V}.
\end{split}
\end{equation}
Across heads, $\{\widetilde{\bm{Y}}^{(j)}\}_{j=1}^{J}$ are concatenated and projected as in \S\ref{sec:pre} to yield $\widetilde{\bm{X}}^{\star}\in\mathbb{R}^{N\times D}$. The $H_pW_p$ patch rows of $\widetilde{\bm{X}}^{\star}$ replace $\bm{v}[h,w]$ in Eq.~\eqref{eq:pre_logits}, yielding Procrustes-aligned patch-text logits gived by:
\begin{equation}
\label{eq:postPA_logits}
\widetilde{\bm{Z}}[h,w,c]=\alpha\,\frac{\widetilde{\bm{v}}[h,w]\cdot \bm{t}_c}{\|\widetilde{\bm{v}}[h,w]\|_2\,\|\bm{t}_c\|_2},\qquad
\alpha=D^{-1/2}.
\end{equation}

This step acts where attention scores are formed. Weighted centering in Eq.~\eqref{eq:center} dampens the influence of high-norm background tokens and the \texttt{CLS} summary, while the orthogonal map in Eq.~\eqref{eq:proc} rotates the key basis toward the query basis without changing local magnitudes. The result is a set of patch features whose inner products better reflect directional agreement in the query subspace, which stabilizes the downstream cosine similarities. The extra cost per head is a compact $d\times d$ SVD and two $N\times d$ multiplications, comparable to the baseline attention.

\subsection{Text-aware Laplacian Propagation}
\label{sec:tlp}

Even with improved geometry, the score field can remain locally noisy in regions with weak evidence or fine structures. A single refinement pass converts these scores into coherent masks by coupling image boundaries with relations induced by the text prototypes. Language serves not only as a classifier but also as a structural prior that governs how labels propagate across neighboring patches.

\begin{table*}[t]
\centering
\setlength{\tabcolsep}{3pt}
\caption{\textbf{Quantitative results of open-vocabulary semantic segmentation.} 
``Extra data" denotes external datasets (\eg, CC3M~\cite{cc3m}, CC12M~\cite{cc12m}, RedCaps~\cite{redcaps}, COCO Captions~\cite{coco_captions,coco}, and ImageNet-1K~\cite{imagenet}), and ``Extra backbone" lists auxiliary models. ``Training-free" indicates no extra training. 
All metrics are mIoU (\%). Best results are highlighted with \colorbox{Best}{\best{bold}}, and second best with \colorbox{Second}{\second{underlined}}.}
\label{tab:ovseg}
\vspace{-0.3cm}
\resizebox{\linewidth}{!}{
\begin{tabular}{l c c c c ccc ccccc c}
\toprule
\multirow{2}{*}{\textbf{Method}} &
\multirow{2}{*}{\textbf{Pub.\,\&\,Year}} &
\multirow{2}{*}{\textbf{Extra data}} &
\multirow{2}{*}{\makecell{\textbf{Extra}\\\textbf{backbone}}} &
\multirow{2}{*}{\makecell{\textbf{Training}\\\textbf{free}}} &
\multicolumn{3}{c}{\textbf{with background}} &
\multicolumn{5}{c}{\textbf{without background}} &
\multirow{2}{*}{\textbf{Avg.}} \\
\cmidrule(lr){6-8}\cmidrule(lr){9-13}
& & & & & V21 & PC60 & Object & V20 & PC59 & Stuff & City & ADE & \\
\midrule
\rowcolor{black!5}
\multicolumn{14}{l}{\textbf{(a) \textit{Trained + Extra data}}}\\
GroupViT~\cite{groupvit}  & CVPR'22 & CC12M+RedCaps & \xmark & \xmark &
50.4 & 18.7 & 27.5 & 79.7 & 23.4 & 15.3 & 11.1 & 9.2  & 29.4 \\
TCL~\cite{tcl}            & CVPR'23 & CC3M+CC12M     & \xmark & \xmark &
51.2 & 24.3 & 30.4 & 77.5 & 30.3 & 19.6 & 23.1 & 14.9 & 33.9 \\
CoDe~\cite{code}          & CVPR'24 & CC3M+RedCaps   & \xmark & \xmark &
57.7 & 30.5 & 32.3 & -   & -   & 23.9 & 28.9 & 17.7 & -   \\
\rowcolor{black!8}
\multicolumn{14}{l}{\textbf{~~~~ \textit{Trained + Extra data \& backkbone}}}\\
SAM-CLIP~\cite{sam-clip}  & CVPR'24 & Merged-41M   & SAMv1 (ViT-B/16) & \xmark &
60.6 & 29.2 & -  & -   & -   & \best{31.5}  & -  & 17.1 & -   \\
CLIP\mbox{-}DINOiser~\cite{dinoisers} & ECCV'24 & ImageNet-1K & DINOv1 (ViT-B/16) & \xmark &
62.1 & 32.4 & 34.8 & 80.9 & 35.9 & 24.6 & 31.7 & \second{20.0} & 40.3 \\
\midrule
\rowcolor{black!5}
\multicolumn{14}{l}{\textbf{(b) \textit{Training-free + Extra data}}}\\
ReCo~\cite{reco}   & NeurIPS'22 & ImageNet-1K     & \xmark & \cmark &
25.1 & 19.9 & 15.7 & 57.7 & 22.3 & 14.8 & 21.6 & 11.2 & 23.5 \\
FOSSIL~\cite{fossil} & WACV'24   & COCO Captions & \xmark & \cmark &
-   & -   & -   & -   & 35.8 & 24.8 & 23.2 & 18.8 & -   \\
\rowcolor{black!8}
\multicolumn{14}{l}{\textbf{~~~~ \textit{Training-free + Extra data \& backkbone}}}\\
FreeDA~\cite{freeda} & CVPR'24   & COCO Captions & DINOv2 (ViT-B/14) & \cmark &
51.7 & 32.6 & 24.4 & 77.1 & 37.1 & 24.9 & 34.0 & 19.5 & 37.7 \\
\midrule
\rowcolor{black!5}
\multicolumn{14}{l}{\textbf{(c) \textit{Training-free + Extra backbone}}}\\
PnP-OVSS~\cite{pnp-ovss} & CVPR'24 & \xmark & BLIP (ViT-L/16) & \cmark &
-   & -   & 36.2 & 51.3 & 28.0 & 17.9 & -   & 14.2 & -   \\
LaVG~\cite{lavg}           & ECCV'24 & \xmark & DINOv1 (ViT-B/8) & \cmark &
62.1 & 31.6 & 34.2 & 82.5 & 34.7 & 23.2 & 26.2 & 15.8 & 38.8 \\
ProxyCLIP$^\ast$~\cite{ProxyCLIP} & ECCV'24 & \xmark & DINOv2$^\dagger$ (ViT-B/14) & \cmark &
    58.6 & 33.8 & \second{37.4} & 83.0 & 37.2 & 25.4 & 33.9 & 19.7 & 41.1 \\
LPOSS$^\ast$~\cite{lposs}  & CVPR'25 & \xmark & DINOv1 (ViT-B/16) & \cmark &
61.1 & \second{34.6} & 33.4 & 78.8 & \second{37.8} & 25.9 & \second{37.3} & \best{21.8} & 41.3 \\
CASS$^\ast$~\cite{cass}  & CVPR'25 & \xmark & DINOv2$^\dagger$ (ViT-B/14) & \cmark &
58.6 & 32.1 & 33.1 & 86.1 & 35.3 & 23.9 & 34.1 & 17.6 & 40.1 \\
CASS$^\ast$~\cite{cass}  & CVPR'25 & \xmark & DINOv3 (ViT-B/16) & \cmark &
\second{62.3} & 34.0 & 36.0 & \best{87.6} & 37.6 & 25.4 & 36.2 & 18.8 & \second{42.2} \\
\midrule
\rowcolor{black!5}
\multicolumn{14}{l}{\textbf{(d) \textit{Training-free + No extra data \& backbone}}}\\
CLIP~\cite{clip}        & ICML'21 & \xmark & \xmark & \cmark &
18.6 & 7.8  & 6.5  & 49.1 & 11.2 & 7.2  & 6.7  & 3.2  & 13.8 \\
MaskCLIP~\cite{maskclip}& ECCV'22 & \xmark & \xmark & \cmark &
38.3 & 23.6 & 20.6 & 74.9 & 26.4 & 16.4 & 12.6 & 9.8  & 27.9 \\
GEM~\cite{gem}          & CVPR'24 & \xmark & \xmark & \cmark &
46.2 & -   & -   & -   & 32.6 & 15.7 & -   & -   & -   \\
CaR~\cite{CaR}          & CVPR'24 & \xmark & \xmark & \cmark &
48.6 & 13.6 & 15.4 & 73.7 & 18.4 & -   & -   & 5.4  & -   \\
CLIPtrase~\cite{cliptrase} & ECCV'24 & \xmark & \xmark & \cmark &
50.9 & 29.9 & \best{43.6} & 81.0 & 33.8 & 22.8 & 21.3 & 16.4 & 32.7 \\
ClearCLIP~\cite{clearclip} & ECCV'24 & \xmark & \xmark & \cmark &
51.8 & 32.6 & 33.0 & 80.9 & 35.9 & 23.9 & 30.0 & 16.7 & 38.1 \\
SCLIP$^\ast$~\cite{sclip}      & ECCV'24 & \xmark & \xmark & \cmark &
59.1 & 30.4 & 30.5 & 80.4 & 34.1 & 22.4 & 32.2 & 16.1 & 38.2 \\
NACLIP$^\ast$~\cite{naclip}    & WACV'25 & \xmark & \xmark & \cmark &
58.9 & 32.2 & 33.2 & 79.7 & 35.2 & 23.3 & 35.5 & 17.4 & 39.4 \\
SFP$^\ast$~\cite{sfp}    & ICCV'25 & \xmark & \xmark & \cmark &
56.8 & 32.3 & 32.1 & 83.4 & 36.0 & 24.0 & 34.1 & 18.1 & 39.6 \\
\midrule
\textbf{PEARL (Ours)} &  & \xmark & \xmark & \cmark &
\best{64.1} & \best{35.1} & 37.3 &
\second{86.9} & \best{38.6} & \second{26.3} & \best{37.6} & 19.4 & \best{43.2} \\
\bottomrule
\end{tabular}
}
\parbox{\linewidth}{\footnotesize
~\emph{Notes:} ``$\ast$" denotes performance reproduced in this work. ``$\dagger$" indicates DINOv2 with registers~\cite{dinov2-reg}. Dashes (-) denote numbers not reported.}
\vspace{-0.8cm}
\end{table*}

Let $\widehat{\widetilde{\bm{Z}}}\in\mathbb{R}^{H\times W\times C}$ be the upsampled logits from Eq.~\eqref{eq:postPA_logits}. They are averaged on a small grid of size $H_g\times W_g$ (adaptive pooling), giving $\bm{Z}_g\in\mathbb{R}^{C\times H_g\times W_g}$. On each grid node $i$ of a 4-connected graph $\mathcal{G}=(\mathcal{V},\mathcal{E})$, set $\bm{p}_i=\mathtt{softmax}\big(\bm{Z}_{g,i}\big)\in\mathbb{R}^{C}$.
Text prototypes define a class similarity that encodes semantic proximity:
\begin{equation}
\label{eq:textsim}
\bm{G}=\mathtt{row\text{-}softmax}\!\big(\bm{T}\bm{T}^\top/\tau_s\big)+\beta \mathbb{I}_C,
\end{equation}
where $\mathtt{row\text{-}softmax}$ applies a softmax independently to each row, $\tau_s>0$ is a temperature, and $\mathbb{I}_C$ is the $C\times C$ identity. After adding the diagonal term, we re-normalize rows to sum to $1$. Confidence at node $i$ combines peak probability with agreement to the text prior:
\begin{equation}
\label{eq:conf}
\begin{split}
    \gamma_i&=\max_{c}p_i(c),\qquad
    u_i=\bm{p}_i^\top \bm{G}\,\bm{p}_i,\\
    \rho_i&=\big(\max\{\gamma_i,\epsilon\}\big)^2\big(1+u_i\big),
\end{split}
\end{equation}
where $\epsilon>0$ avoids vanishing weights. Neighboring nodes $(i,j)\in\mathcal{E}$ receive an image edge strength and a text-consistency gate, which are formulated as follows:
\begin{equation}
\label{eq:edges}
\begin{split}
    b^{\text{img}}_{ij}&=\exp\!\big(-\kappa\,\|\nabla \bm{I}\|_{ij}\big),\\
    g_{ij}&=\mathtt{clip}_{[0,1]}\big(\bm{p}_i^\top \bm{G}\,\bm{p}_j\big),\\
    a_{ij}&=b^{\text{img}}_{ij}\big(1+\lambda\,g_{ij}\big),
\end{split}
\end{equation}
with $\kappa>0$ and $\lambda\ge 0$.
Here $\|\nabla \bm{I}\|_{ij}:=|\tilde{I}_i-\tilde{I}_j|$ denotes the grayscale difference on edge $(i,j)$, and $\tilde{I}=\mathtt{Gray}(\bm{I})$.
Let $\bm{L}$ be the weighted graph Laplacian of $\mathcal{G}$, defined by the weights $\{a_{ij}\}$.
Refined logits on the grid minimize a convex quadratic that balances data trust and smoothness along image- and text-consistent edges, which is given by:
\begin{equation}
\label{eq:energy}
\begin{split}
\mathcal{L}(\bm{F}_g) = & \frac{1}{2}\sum_{i\in\mathcal{V}} \rho_i\,\|\bm{F}_{g,i}-\bm{Z}_{g,i}\|_2^2 \\
& + \frac{\tau}{2}\sum_{(i,j)\in\mathcal{E}} a_{ij}\,\|\bm{F}_{g,i}-\bm{F}_{g,j}\|_2^2,~~ \tau>0.
\end{split}
\end{equation}
The resulting normal equations are:
\begin{equation}
\label{eq:linsys}
\big(\bm{D}_{\rho}+\tau\bm{L}\big)\bm{F}_g=\bm{D}_{\rho}\,\bm{Z}_g,\qquad \bm{D}_{\rho}=\operatorname{diag}\big(\{\rho_i\}\big),
\end{equation}
which are symmetric positive definite when at least one $\rho_i>0$. A small, fixed number of conjugate-gradient iterations suffices on the downsampled grid. The solution is then bilinearly upsampled to $\bm{F}\in\mathbb{R}^{C\times H\times W}$ and the final segmentation is obtained by:
\begin{equation}
\label{eq:finalS}
\bm{S}(\bm{I},\mathcal{Y})[h,w]=\arg\max_{c\in\{1,\ldots,C\}}\,\bm{F}[c,h,w].
\end{equation}

This construction turns language into structure: (\textit{i}) Eq.~\eqref{eq:textsim} captures which classes tend to co-occur or be visually close in the text space, (\textit{ii}) Eq.~\eqref{eq:conf} enhances the influence of reliable nodes that agree with that prior, and (\textit{iii}) Eq.~\eqref{eq:edges} permits propagation mostly along boundaries that are weak in the image and strong in the text sense. Combined with the alignment in \S\ref{sec:pa}, the pipeline first fixes the geometry that produces the scores and then performs one principled propagation step, achieving coherent masks without training and with modest computational overhead.

%% file: sec/4_experiments.tex
\begin{table*}[t]
\centering
\caption{\textbf{Quantitative results of open-vocabulary semantic segmentation using average pixel accuracy (pAcc, \%)}. Rows are grouped by whether an \emph{extra vision backbone} is used.
Best results are highlighted with \colorbox{Best}{\best{bold}}, and second best with \colorbox{Second}{\second{underlined}}.}
\vspace{-0.3cm}
\label{tab:pacc}
\setlength{\tabcolsep}{8.6pt}
\resizebox{\linewidth}{!}{
\begin{tabular}{lccccccccccc}
\toprule
\multirow{2}{*}{\textbf{Method}} &
\multirow{2}{*}{\textbf{Pub.\,\&\,Year}} &
\multirow{2}{*}{\makecell{\textbf{Extra}\\\textbf{backbone}}} &
\multicolumn{3}{c}{\textbf{with background}} &
\multicolumn{5}{c}{\textbf{without background}} &
\multirow{2}{*}{\textbf{Avg.}} \\
\cmidrule(lr){4-6}\cmidrule(lr){7-11}
& & & V21 & PC60 & Object & V20 & PC59 & Stuff & City & ADE & \\
\midrule
\rowcolor{black!5}
\multicolumn{12}{l}{\textbf{Training-free \emph{with} extra backbone}}\\
LaVG~\cite{lavg}            & ECCV'24 & DINOv1 (ViT-B/8)          & \best{89.3} & 48.7 & 74.8 & 91.1 & 58.9 & 39.1 & 68.5 & 37.0 & 63.4 \\
ProxyCLIP$^\ast$~\cite{ProxyCLIP}  & ECCV'24 & DINOv2$^\dagger$ (ViT-B/14) & 86.6 & 52.0 & \best{75.9} & 88.4 & \best{63.4} & \best{43.4} & \second{74.9} & \best{49.1} & 66.7 \\
CASS$^\ast$~\cite{cass}       & CVPR'25 & DINOv2$^\dagger$ (ViT-B/14) & 85.6 & 50.9 & 70.9 & 92.2 & 60.1 & 41.1 & 72.5 & 41.0 & 64.3 \\
CASS$^\ast$~\cite{cass}       & CVPR'25 & DINOv3 (ViT-B/16)         & 88.1 & \best{53.6} & 75.5 & \best{93.8} & 62.3 & 42.0 & \best{75.2} & 45.6 & \second{67.0} \\
\midrule
\rowcolor{black!5}
\multicolumn{12}{l}{\textbf{Training-free \emph{without} extra backbone}}\\
CLIPtrase~\cite{cliptrase}  & ECCV'24 & \xmark & 78.6 & 52.1 & 50.1 & 89.7 & 58.9 & 38.9 & 63.4 & 38.6 & 59.1 \\
SCLIP$^\ast$~\cite{sclip}          & ECCV'24 & \xmark & 87.6 & 49.2 & 74.3 & 91.0 & 58.3 & 38.4 & 72.7 & 38.7 & 63.8 \\
NACLIP$^\ast$~\cite{naclip}        & WACV'25 & \xmark & 87.1 & 51.2 & 75.3 & 89.2 & 59.8 & 39.3 & 71.7 & 45.2 & 64.9 \\
SFP$^\ast$~\cite{sfp}              & ICCV'25 & \xmark & 85.3 & \best{53.6} & 71.3 & 91.9 & 60.7 & 41.3 & 71.7 & 43.8 & 65.0 \\
\midrule
\textbf{PEARL (Ours)}       & --      & \xmark & \second{88.5} & \second{53.5} & \second{75.8} & \second{93.3} & \second{62.9} & \second{42.9} & 73.8 & \second{46.9} & \best{67.2} \\
\bottomrule
\end{tabular}
}
\parbox{\linewidth}{\footnotesize
~\emph{Notes:} ``$\ast$" denotes performance reproduced in this work. ``$\dagger$" indicates DINOv2 with registers~\cite{dinov2-reg}.}
\vspace{-0.6cm}
\end{table*}

\section{Experiments}
\label{sec:exp}

\subsection{Experimental Setup}
\label{sec:exp_settings}

\noindent\textbf{Datasets.}
Evaluation follows the training-free OVSS protocol on eight widely-used standard benchmarks covering settings \emph{with} and \emph{without} an explicit background class.
The \emph{with background} group comprises Pascal VOC 21 (V21) with 21 categories~\cite{pascal_voc}, Pascal Context 60 (PC60) with 60 categories~\cite{pascal_context}, and COCO-Object (Object) with 80 object categories derived from MS-COCO~\cite{coco}.
The \emph{without background} group comprises Pascal VOC 20 (V20) with 20 categories~\cite{pascal_voc}, Pascal Context 59 (PC59) with 59 categories~\cite{pascal_context}, COCO-Stuff (Stuff) with 171 classes~\cite{stuff}, Cityscapes (City) with 19 classes~\cite{Cityscapes}, and ADE20K (ADE) with 150 classes~\cite{ade}.
All results are reported on the official validation splits, using the public class-name lists and the standard ImageNet prompt templates~\cite{clip} (\eg, ``a photo of a \texttt{class}'') without dataset-specific prompt engineering, in line with common practice~\cite{naclip,sclip,sfp}.

\noindent\textbf{Implementation Details.}
PEARL is applied strictly at test time on a frozen vision-language backbone. Unless otherwise stated, we use \textbf{CLIP ViT-B/16} for both the vision and text encoders~\cite{clip}.
Two modular components are enabled: (\textit{i}) \emph{Procrustes alignment} in the last self-attention block, which computes a per-head, per-image orthogonal map to align keys to queries, and (\textit{ii}) \emph{text-aware Laplacian propagation}, which refines the class-logit map on a compact grid by solving a small symmetric positive-definite linear system with a fixed number of conjugate-gradient iterations before the final upsampling. 
By default, the grid size in \S\ref{sec:tlp} is set by $(H_g,W_g)$. we use $(H_g,W_g)=(224,224)$ for City, and $(80,80)$ otherwise.
The larger grid on City prevents oversmoothing of thin structures and fine-grained ``\textit{stuff}" detail in high-resolution scenes. No auxiliary backbones, additional training, or external supervision are used.
Additional experimental details, including model hyperparameter settings, are provided in the appendix.

\noindent\textbf{Evaluation Protocol and Metric.}
Following the prior training-free OVSS protocol~\cite{clearclip,naclip,sclip,cliptrase,sfp}, input images are resized to have a shorter side of $336$ pixels (City uses $560$ due to its higher base resolution). We adopt sliding-window inference with a $224\times224$ crop and stride $112$, following previous works~\cite{naclip,sclip,sfp}. All results are single-scale (no multi-scale, no flips). For fair comparison, \emph{no} DenseCRF~\cite{densecrf} or PAMR~\cite{pamr} refinement is used in any reported number. Performance is measured primarily with mean Intersection-over-Union (\textbf{mIoU}), complemented by pixel accuracy (\textbf{pAcc}) for a broader assessment.
All experiments are run on a single NVIDIA V100 (32\,GB).

\subsection{Comparison with SOTA Methods}
\label{sec:sota}

\noindent\textbf{Quantitative Results (mIoU).}
Table~\ref{tab:ovseg} presents a comprehensive comparison under the training-free OVSS protocol. 
Among methods that \emph{do not} use auxiliary vision backbones, our method, PEARL, delivers the best average mIoU of \textbf{43.2}, clearly surpassing recent strong baselines such as NACLIP~\cite{naclip} (39.4) and SFP~\cite{sfp} (39.6). 
PEARL ranks first on V21~\cite{pascal_voc} (\textbf{64.1}), PC59~\cite{pascal_context} (\textbf{38.6}), and City~\cite{Cityscapes} (\textbf{37.6}), and remains competitive on V20~\cite{pascal_voc} (\textbf{86.9}), where the very best result (87.6) comes from a system that \emph{does} rely on a DINOv3 backbone~\cite{dinov3}.
Compared with training-free approaches that \emph{do} add powerful backbones, PEARL is within striking distance of the best averages (\eg, CASS~\cite{cass} with DINOv3~\cite{dinov3} at 42.2), despite using a single frozen CLIP encoder. 
Two exceptions are worth noting: on Object~\cite{coco} and ADE~\cite{ade}, PEARL trails the very best numbers by a small margin. 
For Object~\cite{coco} (only ``\textit{things}'' classes), methods that explicitly model background cleaning or leverage DINO-style objectness have a slight advantage. PEARL does not use any background-cleaning heuristic. 
For ADE~\cite{ade}, the fine-grained, diverse taxonomy dilutes text-vision agreement at the patch level. we find that generic CLIP prompts sometimes under-specify rare ``\textit{stuff}'' categories, which limits zero-shot matching.

\begin{figure*}[t]
    \centering
    \includegraphics[width=1\linewidth]{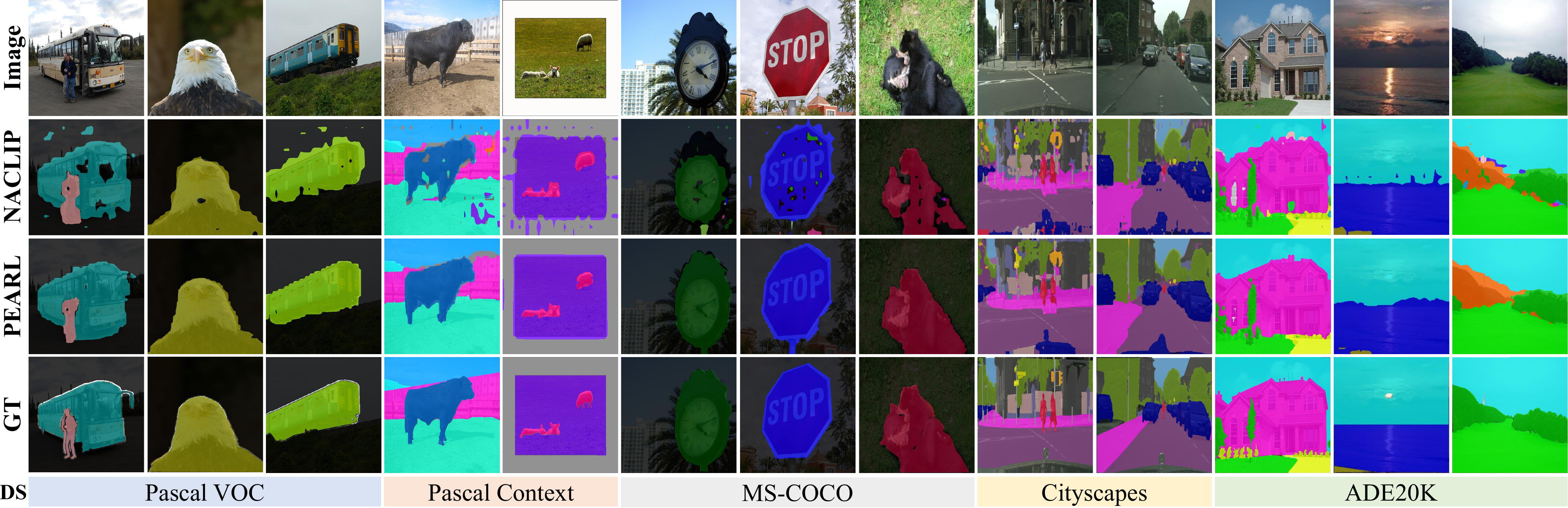}
    \vspace{-0.7cm}
    \caption{\textbf{Qualitative results of open-vocabulary semantic segmentation.} Results are shown on the Pascal VOC~\cite{pascal_voc}, Pascal Context~\cite{pascal_context}, MS-COCO~\cite{coco}, Cityscapes~\cite{cliptrase}, and ADE20K~\cite{ade} datasets (\textbf{DS}), comparing PEARL (\textbf{ours}) with NACLIP~\cite{naclip} and SFP~\cite{sfp}. All methods use CLIP ViT-B/16~\cite{clip}, and no post-processing (\eg, PAMR~\cite{pamr} or DenseCRF~\cite{densecrf}) is applied for a fair comparison.}
    \label{fig:vis}
    \vspace{-0.5cm}
\end{figure*}

\noindent\textbf{Quantitative Results (pAcc).}
Table~\ref{tab:pacc} reports pixel accuracy and compares state-of-the-art training-free OVSS methods.
PEARL achieves the best training-free no extra backbone average of \textbf{67.2}, and is consistently first or second across datasets, especially on V20~\cite{pascal_voc}, PC59~\cite{pascal_context}, and City~\cite{Cityscapes}, indicating that Procrustes alignment stabilizes token geometry for ``\textit{things}'', while the text-aware Laplacian propagation improves coherence in ``\textit{stuff}'' without post-processing.
Where we lag slightly (PC60: 53.5 \textit{vs.}\ SFP~\cite{sfp} 53.6, ADE: 46.9 \textit{vs.}\ ProxyCLIP~\cite{ProxyCLIP} 49.1), the gap stems from the absence of auxiliary DINO-like region grouping, which particularly benefits fine-grained, diverse ``\textit{stuff}'' layouts (\ie, ADE~\cite{ade}) and the broader class set in PC60~\cite{pascal_context}.
PEARL outperforms the top average pAcc (CASS~\cite{cass}\,+\,DINOv3~\cite{dinov3} at 67.0) by \textbf{0.2}, while remaining strictly training-free and using no extra backbone.

\vspace{-0.08cm}
\noindent\textbf{Qualitative Results.}
As shown in Fig.~\ref{fig:vis}, we compare NACLIP~\cite{naclip} and PEARL across scenes from Pascal VOC/Context~\cite{pascal_voc,pascal_context}, MS-COCO~\cite{coco}, Cityscapes~\cite{Cityscapes}, and ADE20K~\cite{ade}.
In this work, our PEARL consistently removes spurious ``\texttt{islands}" and fills missing parts on foreground objects (\eg, \textit{vehicles}, \textit{animals}, \textit{people}).
Interiors are smoother and contain fewer holes than with NACLIP~\cite{naclip}.
Boundaries on large ``\emph{stuff}" (\textit{road}, \textit{sky}, \textit{water}, \textit{facades}) are also cleaner, especially around thin structures such as poles and signs.
These effects reflect the Procrustes alignment, which sharpens token geometry, followed by a single text-aware Laplacian propagation that performs class-conditioned smoothing.
A recurring failure appears in the last ADE20K~\cite{ade} column.
This confusion is common in training-free OVSS~\cite{talk2dino,dino.txt,ProxyCLIP,reme,dih-clip}.
Distant ``\textit{trees}" often form coarse, low-frequency textures that resemble ``\textit{mountain}" patterns, and CLIP~\cite{clip} text prototypes place semantically related classes close in embedding space.
Without task-specific training, richer multi-prompt context, or auxiliary depth/shape cues, the decision can skew toward the more generic terrain label.

\subsection{Ablation Studies}
We ablate the main components of PEARL, \ie, \textit{Procrustes Alignment} (PA) and \textit{Text-aware Laplacian Propagation} (TLP), along with its design and efficiency, reporting all results in mIoU.
See the appendix for more details.

\begin{table}[t]
\vspace{0.11cm}
\centering
\setlength{\tabcolsep}{2pt}
\caption{\textbf{Ablation analysis of key components, \ie, PA (\cf \S\ref{sec:pa}) and TLP (\cf \S\ref{sec:tlp}).} ``BG" means the background class.}
\label{tab:abl_comp}
\vspace{-0.3cm}
\resizebox{\linewidth}{!}{
\begin{tabular}{cc ccc ccccc c c}
\toprule
\multicolumn{2}{c}{\textbf{Component}} &
\multicolumn{3}{c}{\textbf{with BG}} &
\multicolumn{5}{c}{\textbf{without BG}} &
\multirow{2}{*}{\textbf{Avg.}} \\
\cmidrule(lr){3-5}\cmidrule(lr){6-10}
~~~\textbf{PA} & \textbf{TLP} & V21 & PC60 & Object & V20 & PC59 & Stuff & City & ADE & \\
\midrule
~~~\xmark & \xmark & 18.6 & 7.8  & 6.5  & 49.1 & 11.2 & 7.2  & 6.7  & 3.2  & 13.8 \\
~~~\cmark & \xmark & 59.2 & 33.0 & 34.3 & 85.0 & 35.3 & 24.5 & 35.0 & 17.9 & 40.6 \\
~~~\xmark & \cmark & 35.4 & 22.4 & 23.4 & 79.3 & 25.0 & 16.9 & 20.5 & 11.7 & 29.3 \\
\midrule
\rowcolor{black!5}
~~~\cmark & \cmark & \textbf{64.1} & \textbf{35.1} & \textbf{37.3} & \textbf{86.9} & \textbf{38.6} & \textbf{26.3} & \textbf{37.6} & \textbf{19.4} & \textbf{43.2} \\
\bottomrule
\end{tabular}
}
\vspace{-0.1cm}
\end{table}

\begin{table}[t]
\centering
\setlength{\tabcolsep}{2pt}
\caption{\textbf{Ablation analysis of plug-and-play TLP (\cf \S\ref{sec:tlp}).}}
\label{tab:abl_tlp}
\vspace{-0.3cm}
\resizebox{\linewidth}{!}{
\begin{tabular}{l ccc ccccc c}
\toprule
\multirow{2}{*}{\textbf{Method}} &
\multicolumn{3}{c}{\textbf{with BG}} &
\multicolumn{5}{c}{\textbf{without BG}} &
\multirow{2}{*}{\textbf{Avg.}} \\
\cmidrule(lr){2-4}\cmidrule(lr){5-9}
& V21 & PC60 & Object & V20 & PC59 & Stuff & City & ADE & \\
\midrule
SCLIP~\cite{sclip}   &
59.1 & 30.4 & 30.5 & 80.4 & 34.1 & 22.4 & 32.2 & 16.1 & 38.2 \\
\rowcolor{black!5}
+ \textbf{TLP}   &
\up{63.4} & \up{34.5} & \up{36.2} & \up{85.7} & \up{37.5} & \up{25.7} & \up{35.9} & \up{18.7} & \up{42.2} \\
\midrule
NACLIP~\cite{naclip} &
58.9 & 32.2 & 33.2 & 79.7 & 35.2 & 23.3 & 35.5 & 17.4 & 39.4 \\
\rowcolor{black!5}
+ \textbf{TLP}   &
\up{63.3} & \up{34.7} & \up{36.2} & \up{83.9} & \up{38.1} & \up{25.7} & \up{37.9} & \up{18.9} & \up{42.3} \\
\midrule
SFP~\cite{sfp}       &
56.8 & 32.3 & 32.1 & 83.4 & 36.0 & 24.0 & 34.1 & 18.1 & 39.6 \\
\rowcolor{black!5}
+ \textbf{TLP}   &
\up{58.8} & \up{34.1} & \up{35.0} & \up{85.3} & \up{38.1} & \up{25.8} & \up{35.8} & \up{19.4} & \up{41.5} \\
\bottomrule
\end{tabular}
}
\vspace{-0.3cm}
\end{table}

\noindent\textbf{Effect of PA and TLP.}
From Table~\ref{tab:abl_comp}, \emph{vanilla} CLIP~\cite{clip} is the baseline with both modules off. Enabling PA alone raises the average from 13.8 to \textbf{40.6} by fixing the place where attention is formed. PA rotates keys toward queries inside the last attention block so that patch features and text prototypes compare in a better coordinate system. Turning on TLP alone improves the average to \textbf{29.3} by turning noisy scores into smoother and more coherent masks. TLP links pixels that agree in the text space and keeps boundaries where image edges are strong. Activating both delivers the best average of \textbf{43.2} with consistent gains with and without background. Typical improvements include V21 (59.2 $\rightarrow$ \textbf{64.1}), PC59 (35.3 $\rightarrow$ \textbf{38.6}), and City (35.0 $\rightarrow$ \textbf{37.6}).

\begin{table}[t]
\centering
\setlength{\tabcolsep}{2.3pt}
\caption{\textbf{Ablation analysis of using different input image sizes.}}
\label{tab:abl_size}
\vspace{-0.3cm}
\resizebox{\linewidth}{!}{
\begin{tabular}{c ccc ccccc c}
\toprule
\multirow{2}{*}{\textbf{Size}} &
\multicolumn{3}{c}{\textbf{with BG}} &
\multicolumn{5}{c}{\textbf{without BG}} &
\multirow{2}{*}{\textbf{Avg.}} \\
\cmidrule(lr){2-4}\cmidrule(lr){5-9}
& V21 & PC60 & Object & V20 & PC59 & Stuff & City & ADE & \\
\midrule
224px & 59.1 & 33.1 & 36.1 & 86.1 & 36.4 & 24.9 & 29.5 & 17.7 & 40.4 \\
280px & 61.1 & 34.4 & 37.4 & 86.8 & 37.8 & 26.0 & 32.6 & 19.2 & 41.9 \\
\midrule
\rowcolor{black!5}
336px & \textbf{64.1} & \textbf{35.1} & \textbf{37.3} & \textbf{86.9} & \textbf{38.6} & \textbf{26.3} & \textbf{33.9} & \textbf{19.4} & \textbf{42.7} \\
\bottomrule
\end{tabular}
}
\vspace{-0.2cm}
\end{table}

\begin{table}[t]
\centering
\setlength{\tabcolsep}{3.8pt}
\caption{\textbf{Ablation analysis of using different CLIP backbones.}}
\label{tab:abl_clip}
\vspace{-0.3cm}
\resizebox{\linewidth}{!}{
\begin{tabular}{l c c ccc c}
\toprule
\multirow{2}{*}{\textbf{Method}} &
\multirow{2}{*}{\textbf{CLIP}} &
\multicolumn{1}{c}{\textbf{with BG}} &
\multicolumn{3}{c}{\textbf{without BG}} &
\multirow{2}{*}{\textbf{Avg.}} \\
\cmidrule(lr){3-3}\cmidrule(lr){4-6}
& & V21 & PC59 & Stuff & ADE &  \\
\midrule
SCLIP~\cite{sclip}
& \multirow{4}{*}{\rotatebox{0}{ViT-B/32}}
                          & 50.6 & 28.7 & 20.0 & 14.8 & 28.5 \\
NACLIP~\cite{naclip}   &  & 51.1 & 32.4 & 21.2 & 14.9 & 29.9 \\
SFP$^\ast$~\cite{sfp}  &  & 50.1 & 31.9 & 21.3 & 15.7 & 29.8 \\
\rowcolor{black!5}
PEARL & & \textbf{54.5} & \textbf{34.7} & \textbf{23.3} & \textbf{16.6} & \textbf{32.3} \\
\midrule
SCLIP~\cite{sclip}
& \multirow{4}{*}{\rotatebox{0}{ViT-L/14}} 
                           & 44.4 & 25.2 & 17.6 & 10.9 & 24.5 \\
NACLIP~\cite{naclip}    &  & 52.2 & 32.1 & 21.4 & 17.3 & 30.8 \\
SFP$^\ast$~\cite{sfp}   &  & 43.8 & 30.4 & 20.8 & 17.0 & 28.0 \\
\rowcolor{black!5}
PEARL & & \textbf{56.6} & \textbf{36.3} & \textbf{24.8} & \textbf{20.7} & \textbf{34.6} \\
\bottomrule
\end{tabular}
}
\parbox{\linewidth}{\footnotesize
~\emph{Notes:} ``$\ast$" denotes performance reproduced in this work.}
\vspace{-0.9cm}
\end{table}

\noindent\textbf{TLP on Existing Training-Free Methods.}
Table~\ref{tab:abl_tlp} reports the effect of adding plug-and-play TLP to previous methods without changing backbones or prompts. TLP lifts SCLIP~\cite{sclip} from 38.2 to \textbf{42.2}, NACLIP~\cite{naclip} from 39.4 to \textbf{42.3}, and SFP~\cite{sfp} from 39.6 to \textbf{41.5}. Gains are larger on PC59, Stuff, and City, where class-conditioned propagation helps large regions while preserving thin structures.

\noindent\textbf{Input Resolution.}
Table~\ref{tab:abl_size} shows the results when inputs are resized to short sides of 224, 280, and 336 pixels for \emph{all} datasets, including City~\cite{Cityscapes}.
The average increases from 40.4 at 224px to 41.9 at 280px and to \textbf{42.7} at 336px. City improves from 29.5  $\rightarrow$ 32.6 $\rightarrow$ \textbf{33.9}, and ADE from 17.7 $\rightarrow$ 19.2 $\rightarrow$ \textbf{19.4}.
Higher resolution gives more stable patch tokens for PA and clearer local edges for TLP. We adopt 336px by default for a good accuracy-efficiency trade-off and to match prior training-free settings for fair comparison.

\noindent\textbf{Impact of CLIP Backbone.}
As summarized in Table~\ref{tab:abl_clip}, we present the ablation results for different CLIP backbones, \ie, ViT-B/32 and ViT-L/14. PEARL is strongest within each group with averages of \textbf{32.3} (B/32) and \textbf{34.6} (L/14). On ADE, PEARL using ViT-L/14 reaches \textbf{20.7}, which is higher than our ViT-B/16 result reported elsewhere. ADE contains many fine-grained ``\textit{stuff}" regions and diverse scene layouts, and the longer context of L/14 helps TLP connect semantically related areas, while PA maintains the geometry stability. On other datasets, the larger model tends to mix tokens more strongly and weakens local contrast at the patch level, thereby reducing the benefit of PA and making it harder to preserve boundaries. Across all benchmarks, ViT-B/16 remains the most reliable overall choice, as it effectively balances locality and global semantics.
At the same time, ViT-L/14 can be preferable on scenes dominated by broad ``\textit{stuff}" regions such as ADE.

\begin{figure}[t]
\centering
    \begin{subfigure}{1\linewidth}
        \includegraphics[width=\linewidth]{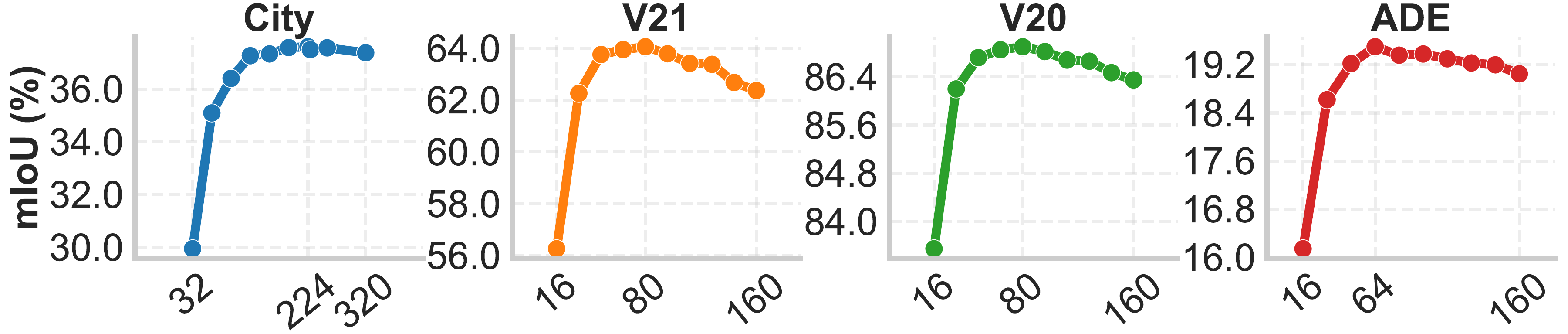}
        \vspace{-0.55cm}
        \caption{Grid Size}
        \label{fig:abl_grid}
    \end{subfigure} 
    \begin{subfigure}{1\linewidth}
        \includegraphics[width=\linewidth]{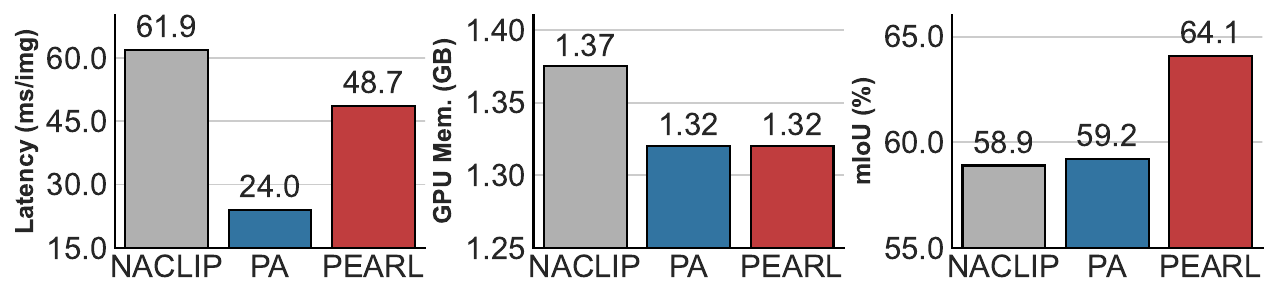}
        \vspace{-0.55cm}
        \caption{Efficiency}
        \label{fig:abl_efficiency}
    \end{subfigure}
    \vspace{-0.8cm}
\caption{\textbf{Ablation analysis of (a) grid size and (b) efficiency.}}
\label{fig:abl_combined}
\vspace{-0.5cm}
\end{figure}

\noindent\textbf{Grid Size \& Efficiency.}
Fig.~\ref{fig:abl_grid} shows mIoU \textit{vs}. grid size: City peaks at 224, V21/V20 about 80, and ADE at 64.
Both overly coarse and fine grids hurt accuracy or efficiency, so we set \textbf{224} for City and \textbf{80} for all other datasets to achieve a fair and simple setup.
As shown in Fig.~\ref{fig:abl_efficiency}, on V21 under this setting, PA already surpasses NACLIP, and adding TLP (PEARL) further improves mIoU to \textbf{64.1}, while reducing memory from 1.37 to 1.32 GB and latency from 61.9 to 48.7 ms/img, yielding the best accuracy-efficiency trade-off.

%% file: sec/5_conclusion.tex
\section{Conclusion}

We present PEARL, a training-free framework for open-vocabulary semantic segmentation that follows an \textit{align-then-propagate} recipe on a frozen vision-language backbone. The first step applies an orthogonal Procrustes alignment within the last self-attention block to align keys with the query subspace after weighted centering, thereby stabilizing token geometry for patch-text matching. The second step performs text-aware Laplacian propagation on a compact grid, where text prototypes provide both a confidence cue and a gate on neighbor links while image edges guide boundaries. Both methods are closed-form, parameter-free, and inexpensive, making the pipeline plug-and-play with CLIP-style inference. Across standard OVSS benchmarks, PEARL consistently improves coherence on small objects and preserves large ``\textit{stuff}" regions under a unified evaluation protocol, without auxiliary backbones or training.

\noindent\textbf{Limitation and Future Work.} Performance depends on prompt quality and label names, very low-contrast boundaries remain challenging, the grid size trades detail for cost, and the method is not instance-aware. These are common constraints in training-free OVSS and motivate future work on prompt calibration, adaptive grids, and instance cues.

{\small{\noindent\textbf{Acknowledgement.}
This work was supported by the National Natural Science Foundation of China (No. 62472222), Natural Science Foundation of Jiangsu Province (No. BK20240080), and in part by the Institute of Information \& communications Technology Planning \& Evaluation (IITP) grant funded by the Korea government (MSIT) (No. RS-2024-00337489).}}

%% file: sec/6_appendix.tex
\clearpage
\setcounter{page}{1}
\maketitlesupplementary

\appendix
\renewcommand\thefigure{\Alph{section}\arabic{figure}}
\renewcommand\thetable{\Alph{section}\arabic{table}}

\section*{Appendix} \label{sec:appendix}

This appendix presents further model settings (\S\ref{sec:supp_set}), ablation studies (\S\ref{sec:supp_abl}), quantitative (\S\ref{sec:supp_quant}), and qualitative results (\S\ref{sec:supp_qual}).

\section{More Model Settings}
\label{sec:supp_set}

\noindent\textbf{Hyperparameter setting.}
For fair and consistent evaluation, we use a unified hyperparameter configuration for all datasets without dataset-specific tuning. The detailed settings are summarized in Table~\ref{tab:hyperparameters}.

\begin{table}[h]
\vspace{0.17cm}
\centering
\caption{\textbf{Fixed hyperparameter setting used for all datasets.}}
\label{tab:hyperparameters}
\vspace{-0.1cm}
\setlength{\tabcolsep}{12pt}
\resizebox{\linewidth}{!}{
\begin{tabular}{ccccccc}
\toprule
\textbf{Config} & $\tau_s$ & $\beta$ & $\epsilon$ & $\kappa$ & $\lambda$ & $\tau$ \\
\midrule
\textbf{Value} & 0.5 & 10 & $10^{-6}$ & 5 & 1 & 1 \\
\bottomrule
\end{tabular}}
\end{table}

\section{More Ablation Studies}
\label{sec:supp_abl}

\begin{figure*}[t]
    \centering
    \includegraphics[width=0.7\linewidth]{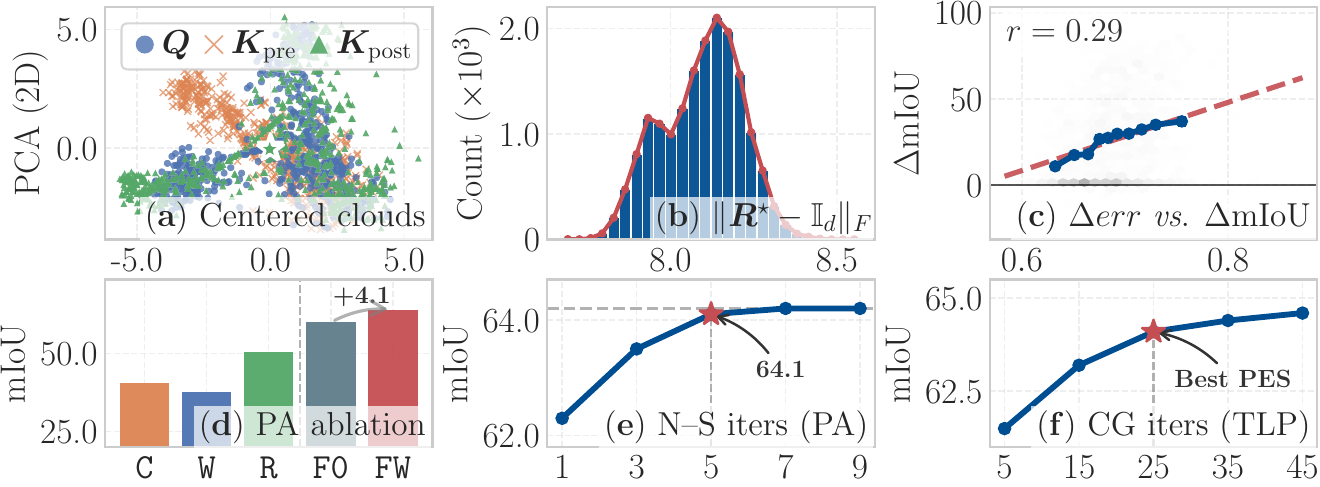}
    \vspace{-0.3cm}
    \caption{\textbf{Diagnostics of Procrustes Alignment on V21 \cite{pascal_voc}.}
    (a) PCA projection of centered queries and keys before and after PA.
    (b) Distribution of the per-(image, head) rotation magnitude $\|\bm{R}^{\star}-\bm{I}_d\|_F$.
    (c) Correlation between alignment-error reduction $\Delta err$ and mIoU gain $\Delta\mathrm{mIoU}$.
    (d) Component ablation of centering (\texttt{C}), whitening (\texttt{W}), rotation (\texttt{R}), full PA without weights (\texttt{F0}), and full weighted PA (\texttt{FW}).
    (e) Stability of the Newton-Schulz iterations used in PA.
    (f) Stability of the conjugate-gradient iterations used in TLP.}
    \label{fig:pa_diag}
\end{figure*}

\noindent\textbf{Alignment Objective.}
We further analyze the behavior of Procrustes Alignment (PA) on V21~\cite{pascal_voc}. PA is applied only to the \emph{last} self-attention layer, where patch-text logits are formed, and aligns the key basis to the query basis ($\bm{K}\!\rightarrow\!\bm{Q}$) to correct token geometry with minimal disturbance to the original similarity structure. As an orthogonal Procrustes map, $\bm{R}^{\star}$ is the minimum-change, inner-product-preserving transformation under an orthogonality constraint, which explains why simpler variants such as centering only, whitening, or global rotation are consistently inferior. As shown in Fig.~\ref{fig:pa_diag}(a), PA makes the centered query/key clouds substantially better aligned in the projected space. Fig.~\ref{fig:pa_diag}(b) shows that the learned rotations have non-trivial yet well-behaved magnitudes across image-head pairs, while Fig.~\ref{fig:pa_diag}(c) indicates a positive correlation between alignment-error reduction and mIoU improvement. The component ablation in Fig.~\ref{fig:pa_diag}(d) further confirms that the full weighted formulation performs best, improving over its unweighted counterpart by 4.1 mIoU. Finally, Fig.~\ref{fig:pa_diag}(e)-(f), together with Tables~\ref{tab:abl_solver} and \ref{tab:abl_solver_time}, show that the iterative \textit{Newton-Schulz} (N-S) solver is stable in practice: it matches SVD in accuracy while being substantially faster, and both the N-S iterations and the conjugate-gradient (CG) iterations exhibit clear performance plateaus, motivating the default settings used in all experiments.

\begin{table}[t]
\centering
\setlength{\tabcolsep}{2.3pt}
\caption{\textbf{Ablation analysis of different alignment solvers (\cf \S\ref{sec:pa}).} ``N-S" denotes the Newton-Schulz iterative algorithm.}
\label{tab:abl_solver}
\vspace{-0.1cm}
\resizebox{\linewidth}{!}{
\begin{tabular}{c ccc ccccc c}
\toprule
\multirow{2}{*}{\textbf{Solver}} &
\multicolumn{3}{c}{\textbf{with BG}} &
\multicolumn{5}{c}{\textbf{without BG}} &
\multirow{2}{*}{\textbf{Avg.}} \\
\cmidrule(lr){2-4}\cmidrule(lr){5-9}
& V21 & PC60 & Object & V20 & PC59 & Stuff & City & ADE & \\
\midrule
SVD & \textbf{64.2} & \textbf{35.2} & 37.2 & 86.7 & \textbf{38.6} & \textbf{26.3} & \textbf{37.9} & 19.2 & \textbf{43.2} \\
\rowcolor{black!5}
N-S & 64.1 & 35.1 & \textbf{37.3} & \textbf{86.9} & \textbf{38.6} & \textbf{26.3} & 37.6 & \textbf{19.4} & \textbf{43.2} \\
\bottomrule
\end{tabular}
}
\vspace{-0.23cm}
\end{table}

\begin{table}[t]
\centering
\setlength{\tabcolsep}{1.5pt}
\caption{\textbf{Inference latency (ms/img) comparison of alignment solvers.} ``N-S" denotes the Newton-Schulz iterative algorithm.}
\label{tab:abl_solver_time}
\vspace{-0.1cm}
\resizebox{\linewidth}{!}{
\begin{tabular}{c ccc ccccc c}
\toprule
\multirow{2}{*}{\textbf{Latency}} &
\multicolumn{3}{c}{\textbf{with BG}} &
\multicolumn{5}{c}{\textbf{without BG}} &
\multirow{2}{*}{\textbf{Avg.}} \\
\cmidrule(lr){2-4}\cmidrule(lr){5-9}
& V21 & PC60 & Object & V20 & PC59 & Stuff & City & ADE & \\
\midrule
SVD & 60.9 & 199.7 & 239.3 & 59.6 & 198.8 & 212.1 & 975.9 & 192.3 & 267.3 \\
\rowcolor{black!5}
N-S & \textbf{48.7} & \textbf{111.7} & \textbf{149.4} & \textbf{47.1} & \textbf{111.2} & \textbf{120.7} & \textbf{498.5} & \textbf{115.4} & \textbf{150.3} \\
\bottomrule
\end{tabular}
}
\vspace{-0.23cm}
\end{table}

\noindent\textbf{Alignment Solver.}
Consistent with the trend in Fig.~\ref{fig:pa_diag}(e), Table~\ref{tab:abl_solver} compares SVD and the N-S iterative solver inside our Procrustes alignment. Both solvers achieve the same average mIoU (\textbf{43.2}), and the per-dataset differences are within 0.3 points: SVD is slightly better on V21/PC60/City, while N-S is slightly better on Object/V20/ADE, indicating that the choice of solver has a negligible impact on accuracy. In terms of efficiency, Table~\ref{tab:abl_solver_time} further shows that N-S consistently yields lower inference latency across all datasets (\eg, \textbf{48.7} \vs. 60.9 ms/img on V21 and \textbf{498.5} \vs. 975.9 ms/img on City), reducing the average latency from 267.3 to \textbf{150.3} ms/img. This speedup is possible because our Procrustes module only requires the orthogonal factor of the SVD of a small $C{\times}C$ matrix: this factor coincides with the orthogonal polar factor $\bm{M}(\bm{M}^{\top}\bm{M})^{-1/2}$, where the inverse square root $(\bm{M}^{\top}\bm{M})^{-1/2}$ can be efficiently approximated by the N-S iteration (\cf \S\ref{sec:pa}) using only matrix multiplications on the GPU. Therefore, we adopt the SVD-free N-S variant as our default solver, which preserves accuracy while substantially improving efficiency.

\noindent\textbf{Key-Key Self-Correlation.}
As shown in Table~\ref{tab:abl_kk}, we evaluate the effect of adding a key-key self-correlation term to our Procrustes Alignment (\cf \S\ref{sec:pa}) and enabling this term (\textit{w/}) improves results on all datasets. In the ``with BG" group, it brings gains of +0.9, +0.4, and +0.5 mIoU on V21~\cite{pascal_voc}, PC60~\cite{pascal_context}, and Object~\cite{coco}, respectively. In the ``without BG" group, it yields +0.5, +0.3, +0.2, +3.0, and +0.7 mIoU on V20~\cite{pascal_voc}, PC59~\cite{pascal_context}, Stuff~\cite{stuff}, City~\cite{Cityscapes}, and ADE~\cite{ade}. Overall, the average mIoU increases from 42.4 to \textbf{43.2} (\up{+0.8}), with no degradation on any dataset and a particularly notable boost on City (\up{+3.0}), where long-range dependencies and cluttered scenes are common.
In our implementation, Procrustes Alignment first recenters queries and keys using the token weights $\pi_n$ in Eq.~\eqref{eq:center} and solves the orthogonal Procrustes problem in Eq.~\eqref{eq:proc}, either via SVD or via an SVD-free Newton-Schulz approximation of the polar factor, to obtain an orthogonal map $\bm{R}^\star$ that aligns the key basis to the query basis. Aligned attention scores are then computed as in Eq.~\eqref{eq:aligned_attn}, and we add a lightweight key-key term constructed from the centered keys, $\bm{K}_c\bm{K}_c^\top$, scaled by a factor $\alpha=d^{-1/2}$. 
Geometrically, the Procrustes term aligns the cross-covariances ($\bm{K}_c^\top\bm{Q}_c$) at the first order. Meanwhile, the key-key Gram matrix captures the self-correlation of $\bm{K}_c$, serving as a second-order regularizer for the attention kernel.
Since $\bm{K}_c$ is already debiased by weighted centering (which suppresses dominant background and \texttt{CLS} tokens), this self-correlation term reinforces coherent foreground regions while dampening isolated noise. Consequently, the attention mechanism merges query alignment with the internal structure of the key space. This yields more stable token interactions and drives the segmentation improvements shown in Table~\ref{tab:abl_kk}.

\noindent\textbf{Impact of Conjugate-Gradient Iterations.}
Table~\ref{tab:abl_cg} and Fig.~\ref{fig:pa_diag}(f) ablate the number of CG iterations used to solve Eq.~\eqref{eq:linsys} in our Text-aware Laplacian Propagation (TLP) on V21~\cite{pascal_voc}.
To balance segmentation quality and efficiency, we define a \emph{Precision-Efficiency Score} (PES) that averages normalized mIoU, pAcc, Latency and GPU Memory for each setting.
For each metric $q \in \{\mathrm{mIoU}, \mathrm{pAcc}\}$ and
$r \in \{\mathrm{Latency}, \mathrm{Memory}\}$, let
$q^{\max}=\max_j q_j$, $q^{\min}=\min_j q_j$ and
$r^{\max}=\max_j r_j$, $r^{\min}=\min_j r_j$.
The normalized scores are represented as follows:
\begin{align}
q_k^{\mathrm{norm}} &=
\begin{cases}
\dfrac{q_k - q^{\min}}{\,q^{\max} - q^{\min}\,}, & q^{\max} > q^{\min},\\[3pt]
1, & \text{otherwise},
\end{cases}
\label{eq:metric_norm_pos} \\[3pt]
r_k^{\mathrm{norm}} &=
\begin{cases}
\dfrac{r^{\max} - r_k}{\,r^{\max} - r^{\min}\,}, & r^{\max} > r^{\min}.\\[3pt]
1, & \text{otherwise}.
\end{cases}
\label{eq:metric_norm_neg}
\end{align}
Here, mIoU and pAcc are ``\textit{the higher the better}'', while Latency and GPU Memory are flipped so that lower cost yields higher normalized scores.
For this ablation, GPU Memory is constant across $k$, so its normalized term is identical for all rows and does not affect the ranking.
The overall \emph{Precision--Efficiency Score} is:
\begin{equation}
\begin{split}
\mathrm{PES}_k =& \frac{1}{4}\big(
\mathrm{mIoU}_k^{\mathrm{norm}} +
\mathrm{pAcc}_k^{\mathrm{norm}} \\
& + \mathrm{Latency}_k^{\mathrm{norm}} +
\mathrm{Memory}_k^{\mathrm{norm}}
\big).
\end{split}
\label{eq:pes}
\end{equation}

On V21, PES peaks at CG=25, yielding clearly higher mIoU and pAcc compared to 5 or 15 iterations. However, further increasing CG to 35 or 45 brings only marginal accuracy gains while incurring noticeably higher latency, thus reducing the overall PES. We therefore fix the number of CG iterations to 25 for all experiments, as this provides an optimal trade-off between precision and efficiency.

\begin{table}[t]
\centering
\setlength{\tabcolsep}{2.0pt}
\caption{\textbf{Ablation analysis of a key-key self-correlation term.}}
\label{tab:abl_kk}
\vspace{-0.1cm}
\resizebox{\linewidth}{!}{
\begin{tabular}{c ccc ccccc c}
\toprule
\multirow{2}{*}{\textbf{key-key}} &
\multicolumn{3}{c}{\textbf{with BG}} &
\multicolumn{5}{c}{\textbf{without BG}} &
\multirow{2}{*}{\textbf{Avg.}} \\
\cmidrule(lr){2-4}\cmidrule(lr){5-9}
& V21 & PC60 & Object & V20 & PC59 & Stuff & City & ADE & \\
\midrule
\textit{w/o} & 63.2 & 34.7 & 36.8 & 86.4 & 38.3 &26.1 & 34.6 & 18.7 & 42.4 \\
\rowcolor{black!5}
\textit{w/} & \textbf{64.1} & \textbf{35.1} & \textbf{37.3} & \textbf{86.9} & \textbf{38.6} & \textbf{26.3} & \textbf{37.6} & \textbf{19.4} & \textbf{43.2} \\
\bottomrule
\end{tabular}
}
\vspace{-0.23cm}
\end{table}

\begin{table}[t]
\centering
\setlength{\tabcolsep}{2.6pt}
\caption{\textbf{Ablation analysis of CG iterations on V21~\cite{pascal_voc} (\cf \S\ref{sec:tlp}.} ``PES" denotes the precision-efficiency score.}
\label{tab:abl_cg}
\vspace{-0.1cm}
\resizebox{\linewidth}{!}{
\begin{tabular}{c cc cc c}
\toprule
\multirow{2}{*}{\textbf{Iteration}} &
\multicolumn{2}{c}{\textbf{Precision}} &
\multicolumn{2}{c}{\textbf{Efficiency}} &
\multirow{2}{*}{\textbf{PES}} \\
\cmidrule(lr){2-3}\cmidrule(lr){4-5}
& mIoU & pAcc & Latency (ms/img) & Memory (GB) & \\
\midrule
5  & 61.5 & 87.4 & \textbf{31.7} & 1.32 & 0.50 \\
15 & 63.2 & 88.2 & 40.5 & 1.32 & 0.73 \\
\rowcolor{black!5}
25 & 64.1 & 88.5 & 48.7 & 1.32 & \textbf{0.80} \\
35 & 64.4 & \textbf{88.7} & 58.5 & 1.32 & 0.79 \\
45 & \textbf{64.6} & \textbf{88.7} & 66.5 & 1.32 & 0.75 \\
\bottomrule
\end{tabular}
}
\vspace{-0.23cm}
\end{table}

\begin{table*}[t]
\centering
\setlength{\tabcolsep}{3pt}
\caption{\textbf{Quantitative results of open-vocabulary semantic segmentation.} 
``Extra data" denotes external datasets (\eg, CC3M~\cite{cc3m}, CC12M~\cite{cc12m}, RedCaps~\cite{redcaps}, COCO Captions~\cite{coco_captions,coco}, and ImageNet-1K~\cite{imagenet}), and ``Extra backbone" lists auxiliary models.
``Training-free" indicates no extra training. 
We evaluate prior methods using their default post-processing (official or re-implemented), while reporting PEARL without any post-processing.
All metrics are mIoU (\%). Best results are highlighted with \colorbox{Best}{\best{bold}}, and second best with \colorbox{Second}{\second{underlined}}.}
\label{tab:supp_ovseg}
\vspace{-0.1cm}
\resizebox{\linewidth}{!}{
\begin{tabular}{l c c c c ccc ccccc c}
\toprule
\multirow{2}{*}{\textbf{Method}} &
\multirow{2}{*}{\textbf{Pub.\,\&\,Year}} &
\multirow{2}{*}{\textbf{Extra data}} &
\multirow{2}{*}{\makecell{\textbf{Extra}\\\textbf{backbone}}} &
\multirow{2}{*}{\makecell{\textbf{Training}\\\textbf{free}}} &
\multicolumn{3}{c}{\textbf{with background}} &
\multicolumn{5}{c}{\textbf{without background}} &
\multirow{2}{*}{\textbf{Avg.}} \\
\cmidrule(lr){6-8}\cmidrule(lr){9-13}
& & & & & V21 & PC60 & Object & V20 & PC59 & Stuff & City & ADE & \\
\midrule
\rowcolor{black!5}
\multicolumn{14}{l}{\textbf{\textit{w/ Mask Refinement}}}\\
GroupViT~\cite{groupvit}  & CVPR'22 & CC12M+RedCaps & \xmark & \xmark &
51.1 & 19.0 & 27.9 & 81.5 & 23.8 & 15.4 & 11.6 & 9.4  & 30.0 \\
TCL~\cite{tcl}            & CVPR'23 & CC3M+CC12M     & \xmark & \xmark &
55.0 & 30.4 & 31.6 & 83.2 & 33.9 & 22.4 & 24.0 & 17.1 & 37.2 \\
CLIP\mbox{-}DINOiser~\cite{dinoisers} & ECCV'24 & ImageNet-1K & DINOv1 (ViT-B/16) & \xmark &
\best{64.6} & 33.5 & 36.1 & 81.5 & 37.1 & 25.3 & 31.5 & \second{20.6} & 41.3 \\
ReCo~\cite{reco}   & NeurIPS'22 & ImageNet-1K     & \xmark & \cmark &
27.2 & 21.9 & 17.3 & 62.4 & 24.7 & 16.3 & 22.8 & 12.4 & 25.6 \\
FreeDA~\cite{freeda} & CVPR'24   & COCO Captions & DINOv2 (ViT-B/14) & \cmark &
52.0 & \best{35.2} & 25.8 & 79.5 & \best{40.2} & \best{27.1} & 34.4 & \best{20.9} & 39.4 \\
\midrule
LaVG~\cite{lavg}           & ECCV'24 & \xmark & DINOv1 (ViT-B/8) & \cmark &
62.1 & 31.6 & 34.2 & 82.5 & 34.7 & 23.2 & 26.2 & 15.8 & 38.8 \\
ProxyCLIP$^\ast$~\cite{ProxyCLIP} & ECCV'24 & \xmark & DINOv2$^\dagger$ (ViT-B/14) & \cmark &
62.0 & \best{35.2} & \best{38.7} & 83.1 & \second{38.9} & \second{26.6} & 35.4 & 20.3 & 42.5 \\
CASS$^\ast$~\cite{cass}  & CVPR'25 & \xmark & DINOv2$^\dagger$ (ViT-B/14) & \cmark &
58.7 & 32.6 & 33.3 & 86.4 & 35.9 & 24.0 & 34.2 & 17.9 & 40.4 \\
CASS$^\ast$~\cite{cass}  & CVPR'25 & \xmark & DINOv3 (ViT-B/16) & \cmark &
62.5 & 34.5 & 36.2 & \best{87.1} & 38.2 & 25.6 & 37.4 & 18.9 & \second{42.6} \\
\midrule
MaskCLIP~\cite{maskclip}& ECCV'22 & \xmark & \xmark & \cmark &
37.2 & 22.6 & 18.9 & 72.1 & 25.3 & 15.1 & 11.2 & 9.0  & 26.4 \\
SCLIP$^\ast$~\cite{sclip}      & ECCV'24 & \xmark & \xmark & \cmark &
61.7 & 31.5 & 32.1 & 83.5 & 36.1 & 23.9 & 34.1 & 17.8 & 40.1 \\
NACLIP$^\ast$~\cite{naclip}    & WACV'25 & \xmark & \xmark & \cmark &
\second{64.1} & 35.0 & 36.2 & 83.0 & 38.4 & 25.7 & \best{38.3} & 19.1 & 42.5 \\
SFP$^\ast$~\cite{sfp}    & ICCV'25 & \xmark & \xmark & \cmark &
58.8 & 34.3 & 34.9 & 85.1 & 38.3 & 25.9 & 36.0 & 19.4 & 41.6 \\
\midrule
\rowcolor{black!5}
\multicolumn{14}{l}{\textbf{\textit{w/o Mask Refinement}}}\\
\textbf{PEARL (Ours)} &  & \xmark & \xmark & \cmark &
\second{64.1} & \second{35.1} & \second{37.3} &
\second{86.9} & 38.6 & 26.3 & \best{37.6} & 19.4 & \best{43.2} \\
\bottomrule
\end{tabular}
}
\parbox{\linewidth}{\footnotesize
~\emph{Notes:} ``$\ast$" denotes performance reproduced in this work. ``$\dagger$" indicates DINOv2 with registers~\cite{dinov2-reg}.}
\vspace{-0.5cm}
\end{table*}

\section{More Quantitative Results}
\label{sec:supp_quant}

Table~\ref{tab:supp_ovseg} presents a quantitative comparison with recent open-vocabulary semantic segmentation methods.
For all comparison baselines, we keep their default post-processing (\eg, PAMR~\cite{pamr} or DenseCRF~\cite{densecrf}), while PEARL is evaluated with CLIP ViT-B/16 only and \emph{without} any mask refinement.
Even in this setting, PEARL achieves the highest average mIoU of \textbf{43.2\%} across the eight benchmarks.
It surpasses the strongest baseline, CASS~\cite{cass} with DINOv3 (42.6\%), by \textbf{0.6} points and outperforms other training-free CLIP-based methods, such as NACLIP~\cite{naclip} (42.5\%) and SFP~\cite{sfp} (41.6\%).
This demonstrates that our alignment and propagation modules provide clear improvements without relying on stronger backbones or extra training data.

\section{More Qualitative Results}
\label{sec:supp_qual}

\begin{figure*}[t]
    \centering
    \includegraphics[width=1\linewidth]{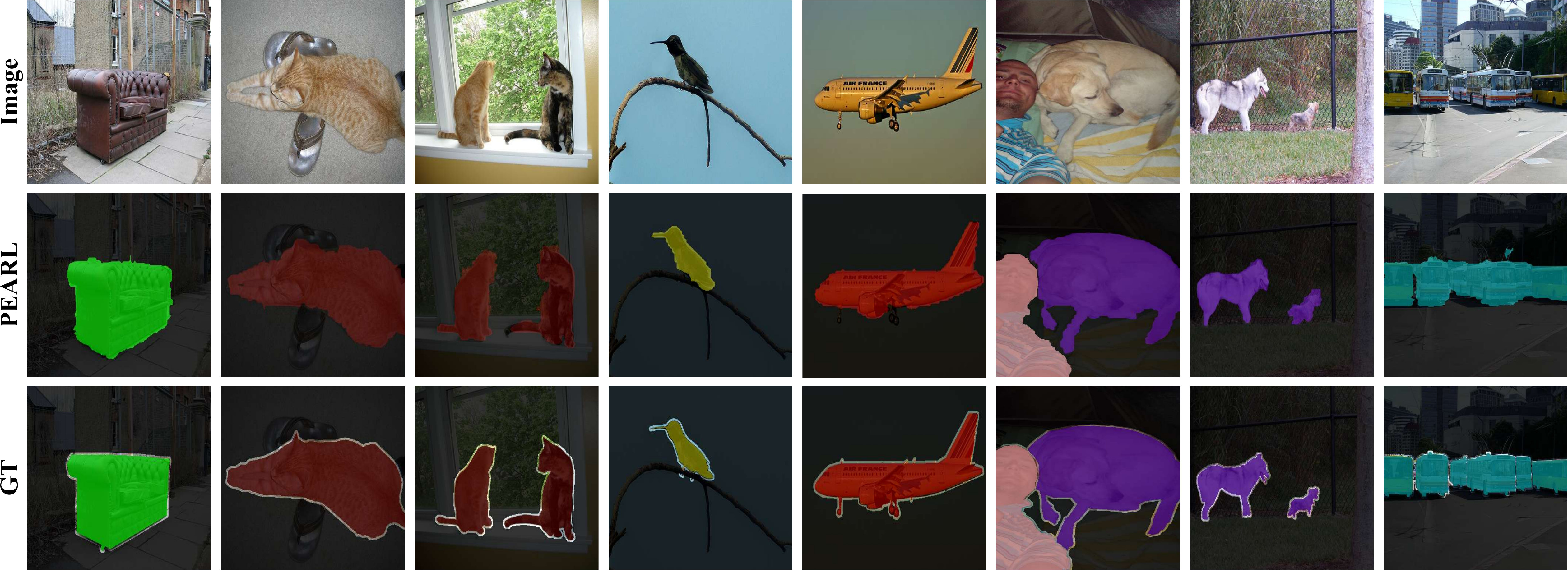}
    \vspace{-0.6cm}
    \caption{\textbf{Qualitative results of open-vocabulary semantic segmentation.} Results are shown on the V21~\cite{pascal_voc} dataset. Our PEARL use CLIP ViT-B/16~\cite{clip}, and no post-processing (\eg, PAMR~\cite{pamr} or DenseCRF~\cite{densecrf}) is applied for a fair comparison.}
    \label{fig:supp_vis_v21}
\end{figure*}

\begin{figure*}[t]
    \centering
    \includegraphics[width=1\linewidth]{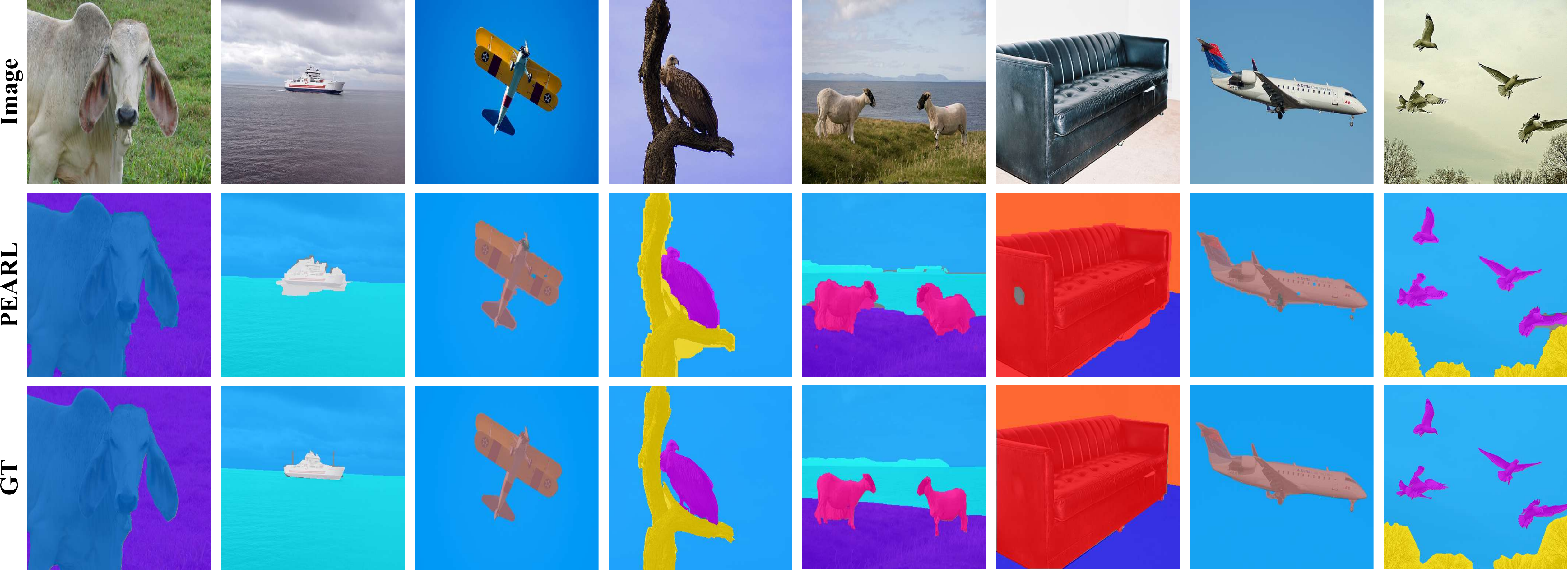}
    \vspace{-0.6cm}
    \caption{\textbf{Qualitative results of open-vocabulary semantic segmentation.} Results are shown on the PC60~\cite{pascal_context} dataset. Our PEARL use CLIP ViT-B/16~\cite{clip}, and no post-processing (\eg, PAMR~\cite{pamr} or DenseCRF~\cite{densecrf}) is applied for a fair comparison.}
    \label{fig:supp_vis_pc60}
\end{figure*}

\begin{figure*}[t]
    \centering
    \includegraphics[width=1\linewidth]{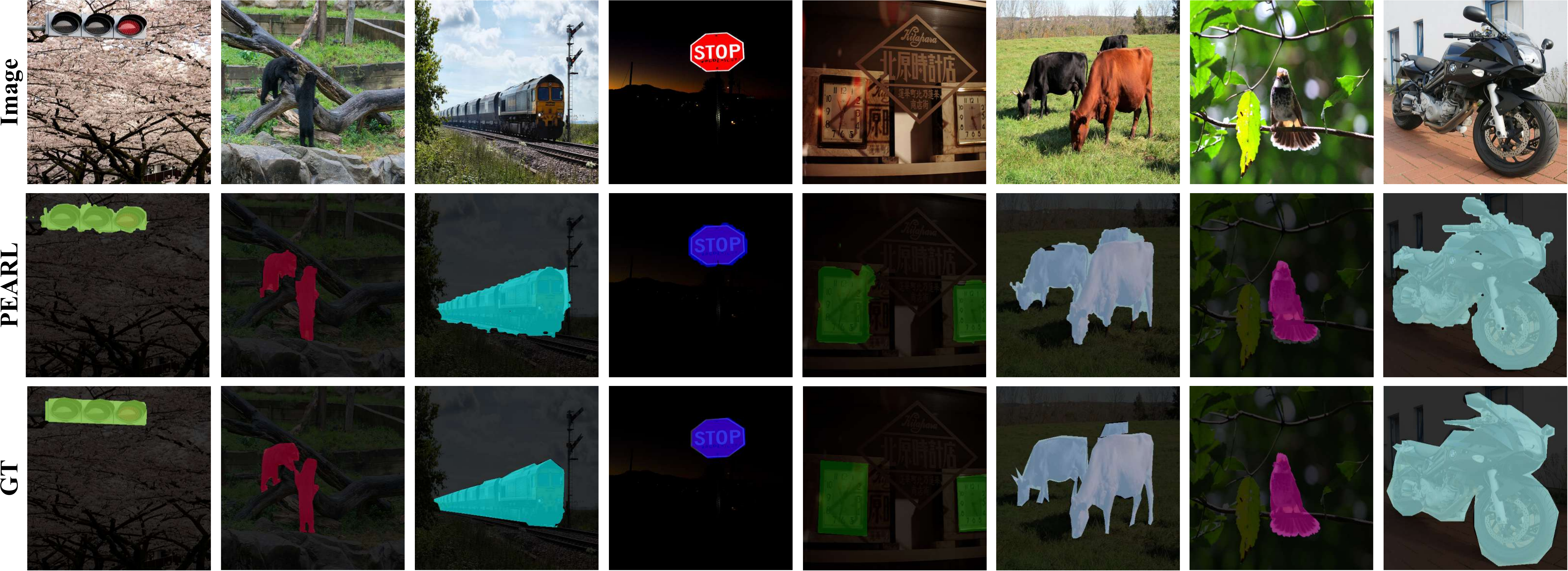}
    \vspace{-0.6cm}
    \caption{\textbf{Qualitative results of open-vocabulary semantic segmentation.} Results are shown on the Object~\cite{coco} dataset. Our PEARL use CLIP ViT-B/16~\cite{clip}, and no post-processing (\eg, PAMR~\cite{pamr} or DenseCRF~\cite{densecrf}) is applied for a fair comparison.}
    \label{fig:supp_vis_object}
\end{figure*}

\begin{figure*}[t]
    \centering
    \includegraphics[width=1\linewidth]{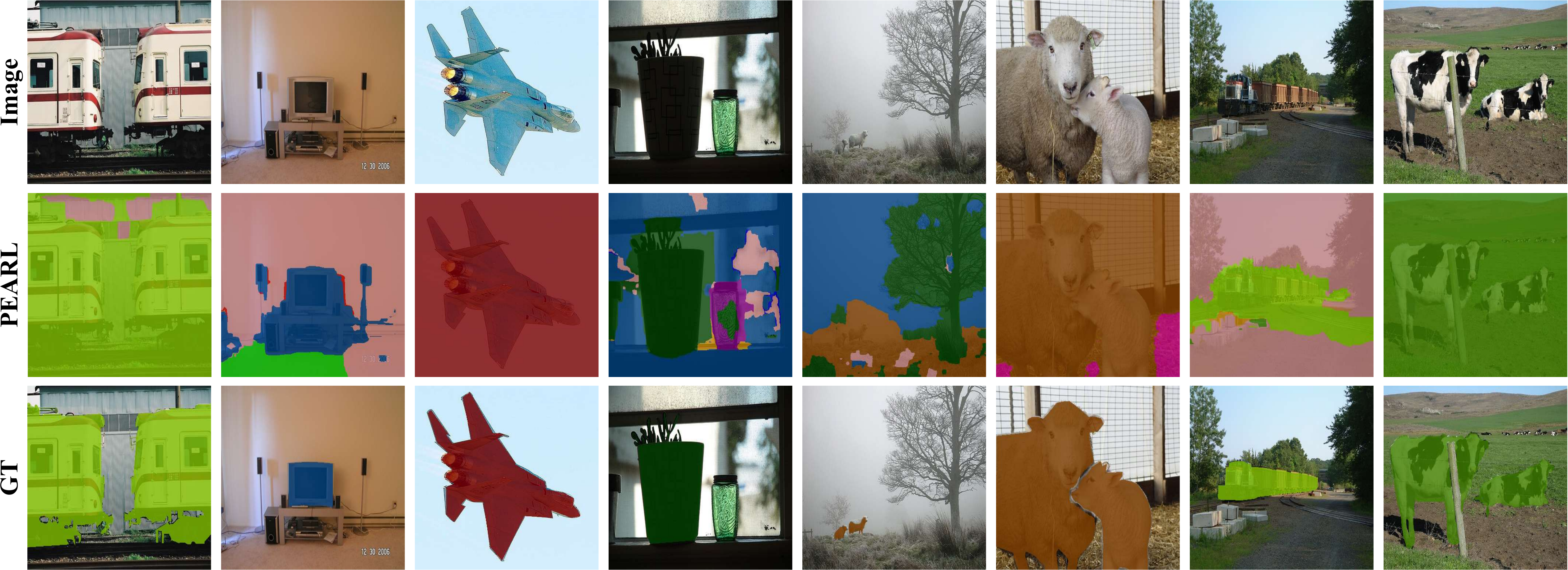}
    \vspace{-0.6cm}
    \caption{\textbf{Qualitative results of open-vocabulary semantic segmentation.} Results are shown on the V20~\cite{pascal_voc} dataset. Our PEARL use CLIP ViT-B/16~\cite{clip}, and no post-processing (\eg, PAMR~\cite{pamr} or DenseCRF~\cite{densecrf}) is applied for a fair comparison.}
    \label{fig:supp_vis_v20}
\end{figure*}

\begin{figure*}[t]
    \centering
    \includegraphics[width=1\linewidth]{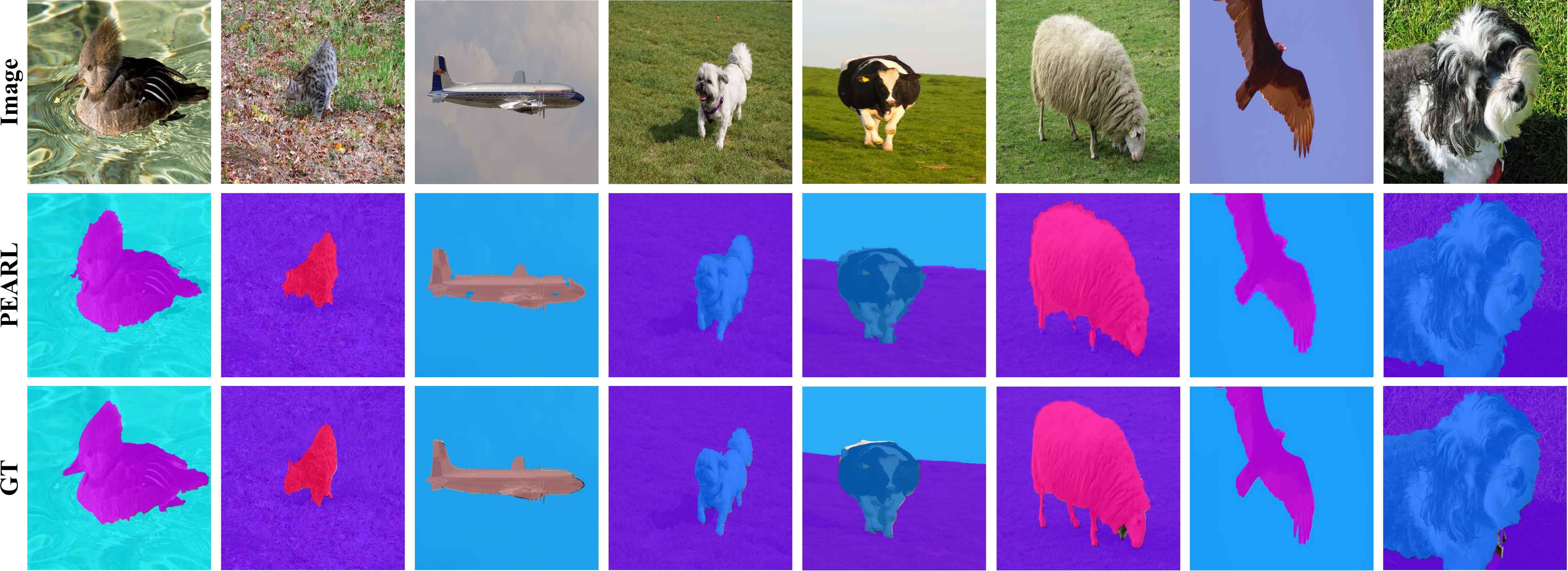}
    \vspace{-0.6cm}
    \caption{\textbf{Qualitative results of open-vocabulary semantic segmentation.} Results are shown on the PC59~\cite{pascal_context} dataset. Our PEARL use CLIP ViT-B/16~\cite{clip}, and no post-processing (\eg, PAMR~\cite{pamr} or DenseCRF~\cite{densecrf}) is applied for a fair comparison.}
    \label{fig:supp_vis_pc56}
\end{figure*}

\begin{figure*}[t]
    \centering
    \includegraphics[width=1\linewidth]{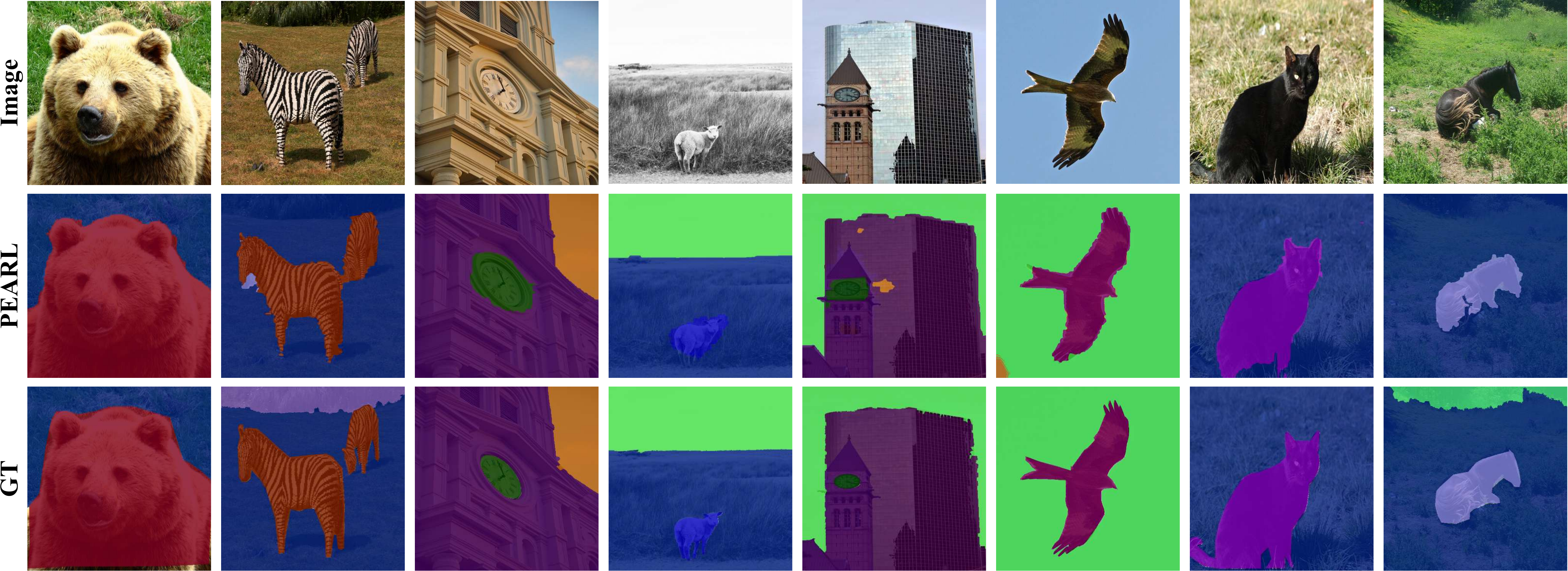}
    \vspace{-0.6cm}
    \caption{\textbf{Qualitative results of open-vocabulary semantic segmentation.} Results are shown on the Stuff~\cite{stuff} dataset. Our PEARL use CLIP ViT-B/16~\cite{clip}, and no post-processing (\eg, PAMR~\cite{pamr} or DenseCRF~\cite{densecrf}) is applied for a fair comparison.}
    \label{fig:supp_vis_stuff}
\end{figure*}

\begin{figure*}[t]
    \centering
    \includegraphics[width=1\linewidth]{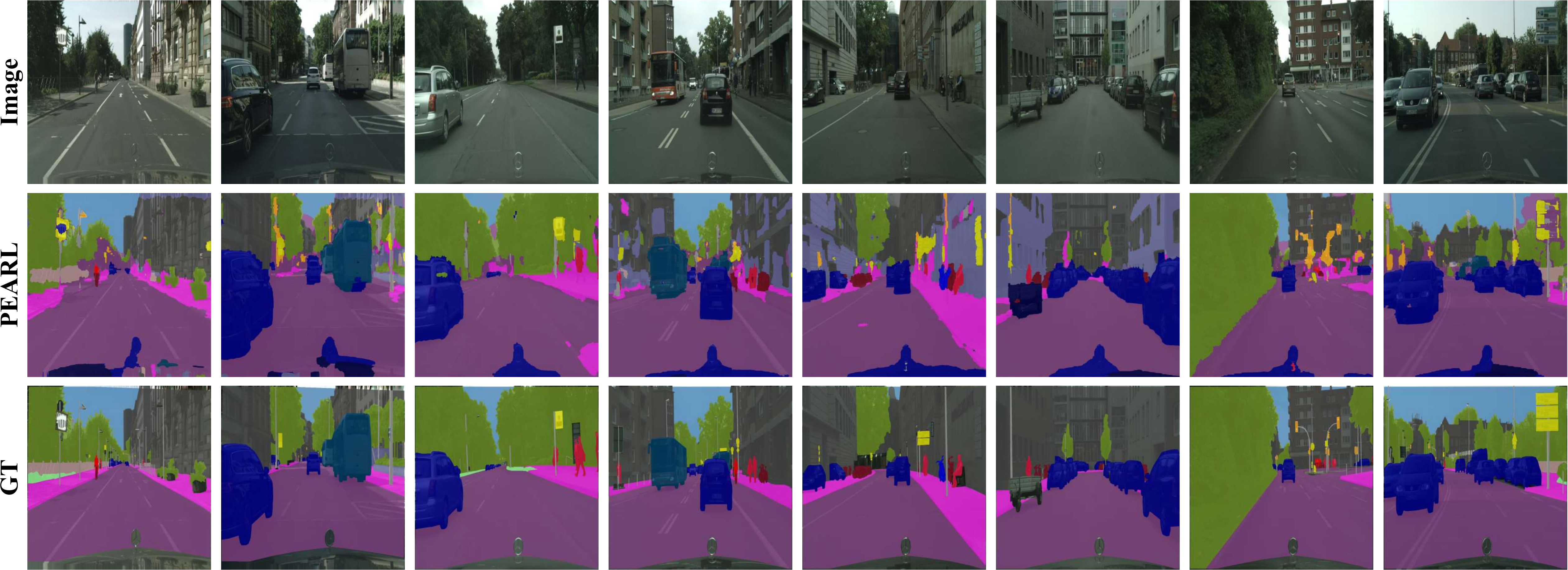}
    \vspace{-0.6cm}
    \caption{\textbf{Qualitative results of open-vocabulary semantic segmentation.} Results are shown on the City~\cite{Cityscapes} dataset. Our PEARL use CLIP ViT-B/16~\cite{clip}, and no post-processing (\eg, PAMR~\cite{pamr} or DenseCRF~\cite{densecrf}) is applied for a fair comparison.}
    \label{fig:supp_vis_city}
\end{figure*}

\begin{figure*}[t]
    \centering
    \includegraphics[width=1\linewidth]{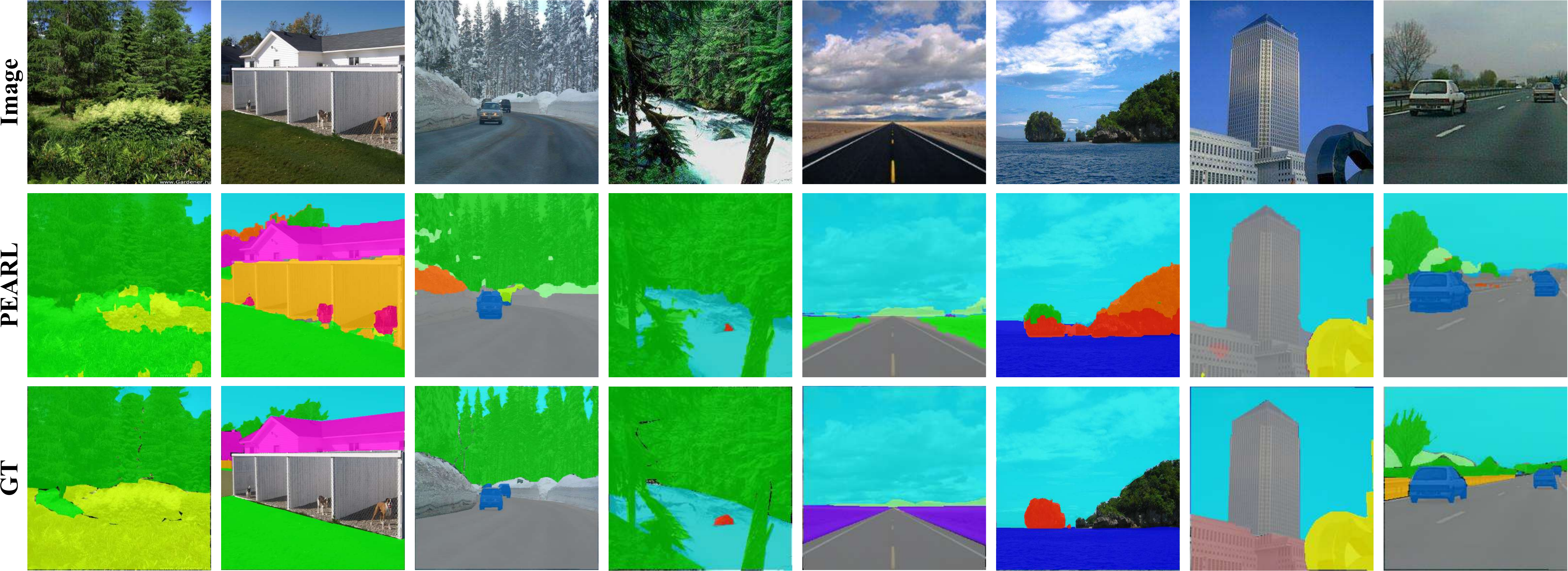}
    \vspace{-0.6cm}
    \caption{\textbf{Qualitative results of open-vocabulary semantic segmentation.} Results are shown on the ADE~\cite{ade} dataset. Our PEARL use CLIP ViT-B/16~\cite{clip}, and no post-processing (\eg, PAMR~\cite{pamr} or DenseCRF~\cite{densecrf}) is applied for a fair comparison.}
    \label{fig:supp_vis_ade}
\end{figure*}

As illustrated in Figs.~\ref{fig:supp_vis_v21}-\ref{fig:supp_vis_ade}, we provide additional qualitative results of our PEARL on all eight benchmarks: Pascal VOC 21 (\textbf{V21})~\cite{pascal_voc}, Pascal Context 60 (\textbf{PC60})~\cite{pascal_context}, COCO-Object (\textbf{Object})~\cite{coco}, Pascal VOC 20 (\textbf{V20})~\cite{pascal_voc}, Pascal Context 59 (\textbf{PC59})~\cite{pascal_context}, COCO-Stuff (\textbf{Stuff})~\cite{stuff}, Cityscapes (\textbf{City})~\cite{Cityscapes}, and ADE20K (\textbf{ADE})~\cite{ade}.
These examples include both successes and typical failure cases.

For each example, we show the input (\textbf{Image}), our prediction (\textbf{PEARL}), and the ground-truth (\textbf{GT}) mask.
All visualizations utilize CLIP ViT-B/16 as the vision backbone, and no post-processing (\eg, PAMR~\cite{pamr} or DenseCRF~\cite{densecrf}) is applied. Therefore, the masks directly reflect the behavior of our training-free pipeline.
On V21/V20~\cite{pascal_voc} and PC60/PC59~\cite{pascal_context}, our PEARL produces accurate object extents and clean boundaries for a wide variety of categories, including animals, vehicles, and artificial objects.
The method remains robust under large appearance changes (\eg, illumination and pose) and complex foreground-background compositions, and it preserves small details such as thin structures and disconnected parts in many cases.
On the Object~\cite{coco} and Stuff~\cite{stuff} datasets, our PEARL can localize both foreground instances and amorphous ``\textit{stuff}" regions, showing that the proposed Procrustes alignment and text-aware propagation generalize well from object-centric images to more cluttered scenes.

For the more challenging City~\cite{Cityscapes} and ADE~\cite{ade} benchmarks, our PEARL still captures the dominant layouts and most large regions (\textit{road, building, sky, vegetation, cars}), but some fine-grained structures and rare categories are not perfectly segmented.
The failure cases in figures mainly fall into several patterns: boundary leakage between adjacent regions, missing or fragmented small objects, and confusion between semantically related classes under cluttered scenes or weak visual evidence.
These failure modes highlight the remaining gap between current open-vocabulary semantic segmentation and fully supervised models on large-scale, high-resolution urban or scene parsing datasets: long-range context, small distant objects, and heavily overlapping classes remain difficult to resolve using frozen backbones and text prompts alone.
We hope that these visualizations will motivate future work on stronger open-vocabulary priors and the better exploitation of geometric and contextual cues in complex, real-world scenes.

